\pdfoutput=1
\documentclass{article}

\usepackage{placeins}
\usepackage[utf8]{inputenc}
\usepackage[T1]{fontenc}
\usepackage[round,authoryear]{natbib}
\usepackage{cleantemplate}
\usepackage{microtype}
\usepackage{charter}

\usepackage{amsmath,amssymb}
\usepackage{graphicx}
\usepackage{booktabs}
\usepackage[font=small,labelfont=bf,hypcap=true]{caption}
\usepackage{array}
\usepackage{algorithm}
\usepackage{algpseudocode}
\ifdefined\pdfsuppressptexinfo \pdfsuppressptexinfo=-1 \fi

\makeatletter
\renewcommand{\@maketitle}{%
  \vbox{%
    \hsize\textwidth
    \linewidth\hsize
    \vskip 0.1in
    {\raggedright\fontsize{17}{20}\selectfont\bfseries\hyphenpenalty=10000\exhyphenpenalty=10000 Comparative study of global pooling for pre-departure ship fuel consumption prediction: Advantages over per-ship and adaptation-based approaches\par}
    \vskip 0.2in
    {\raggedright\normalsize\textbf{Heungjun~Im\,$^{1\dagger}$, Bongmin~Cha\,$^{2\dagger}$, Agung~Nugraha\,$^{1\dagger}$, Donghyun~Kim\,$^{3}$, Seongphil~Kang\,$^{4}$, Minwook~Choi\,$^{4}$, Jeongmin~Park\,$^{4}$, Jihwan~Lee\,$^{1*}$}\par}
    \vskip 0.1in
    {\small
    $^{1}$\,Department of Industrial and Data Engineering, Major in Industrial Data Science and Engineering, Pukyong National University, Busan, 48513, Republic of Korea\\
    $^{2}$\,DX-Solutions Co., Ltd., Busan, Republic of Korea\\
    $^{3}$\,Division of Mechanical, Shipbuilding \& Air Conditioning Engineering, Pukyong National University, Busan, 48513, Republic of Korea\\
    $^{4}$\,LAB021 Co., Ltd., Busan, Republic of Korea\par}
    \vskip 0.4in
  }%
  \begingroup
    \renewcommand{\thefootnote}{}%
    \renewcommand{\@makefnmark}{}%
    \footnotetext{$^{\dagger}$\,These authors contributed equally (co-first authors).}%
    \footnotetext{$^{*}$\,Correspondence: \texttt{jihwan@pknu.ac.kr}}%
  \endgroup
}
\makeatother

\title{Comparative study of global pooling for pre-departure ship fuel consumption prediction: Advantages over per-ship and adaptation-based approaches}
\author{}

\begin{document}

\maketitle

\begin{abstract}
Most fuel oil consumption (FOC) models for voyage planning are per-ship models and thus generalize poorly beyond each vessel's records. Transfer and federated learning have been reported to perform better, but under different protocols that often use inputs unavailable before departure or random data splits. It therefore remains unclear which approach performs best. In this study, we compare 11 per-ship, adaptation-based, and global methods on 21 vessels. To prevent leakage, all methods receive identical inputs known or forecast before departure, excluding engine RPM and shaft power, and are tested on intact voyage legs after all training records of each vessel. Global pooling, in which one model describes each vessel by its ship identity and seven ship specifications, performs best overall in every setting examined. For in-fleet vessels, Global LightGBM reaches an overall MAPE of 8.92\%, against 11.14\% for Per-Ship LightGBM, and outperforms every per-ship and adaptation-based model. Global models of five architectures, including a Transformer, all outperform the per-ship and federated models. Fine-tuning a global model for each vessel, even by low-rank adaptation of the Transformer, adds no accuracy. Without any records of a vessel, the global model reaches a median MAPE of 11.20\% and outperforms a per-ship model trained on that vessel's records for 14 of 21 vessels. Adding its first records to the pooled data is as accurate as vessel-specific adaptation. Ship identity and specifications are largely interchangeable for in-fleet vessels, but only the specifications support zero-shot prediction. Its error grows as a vessel's specifications differ from the fleet's. Per-ship models and vessel-specific adaptation are therefore unnecessary under fair comparison. One global model can serve a whole fleet, including newly delivered vessels with similar specifications.
\end{abstract}


\section{Introduction}
\label{sec:intro}

Voyage planning requires accurate prediction of fuel oil consumption (FOC) before departure. Operators compare the expected fuel consumption of alternative routes and speeds when planning a voyage \citep{i4_survey}. Since 2023, the International Maritime Organization (IMO) has also rated covered ships annually under the Carbon Intensity Indicator (CII) \citep{mepc328}. Because the CII is based on carbon dioxide emissions per unit of transport work, the rating of a vessel depends directly on its fuel consumption. A prediction model for this purpose must use only inputs that are known or forecast before departure, such as planned speed, draft, and weather forecasts. It must also cover every vessel in a fleet, including newly delivered vessels.

The most common approach to data-driven FOC prediction is \emph{per-ship modeling}, in which a separate model is trained for each vessel \citep{i4_survey, b2_comparison, son2025varnn}. Per-ship models have been built with a wide range of methods, from support vector regression of shaft power \citep{kim2020svr} to gradient-boosted trees \citep{handayani2025foc, l4_comprehensive_eval, b2_comparison} and deep neural networks \citep{d5_cnn_lstm, wang2023ga_lstm, b5_transformer_foc}. These per-ship models, however, are trained only on the records of a single vessel and thus lack generalization capability, that is, perform poorly on new routes and on loading or weather conditions that are rare in the vessel's history. \citet{son2025varnn} formalized this problem as a covariate shift between the voyages of a single bulk carrier. Per-ship models also require a sufficient number of records for each vessel \citep{a1_ann_transfer} and thus cannot be trained at all for a newly delivered vessel.

To overcome the limitations of per-ship modeling, multi-ship learning, which uses the records of multiple vessels, has emerged in the literature \citep{a1_ann_transfer, a5_federated, nguyen2023multiship, zhou2026f2vformer}. Most of these studies follow \emph{adaptation-based approaches}, in which knowledge learned from several vessels is adapted to each target vessel with that vessel's own records. In transfer learning, a model pretrained on source vessels is fine-tuned on the target vessel \citep{a1_ann_transfer, a2_transfer_scarcity, zhou2026f2vformer}. In federated learning, a server averages local models trained on individual vessels without collecting their records \citep{mcmahan2017fedavg}. Personalized variants then adapt the averaged model to each vessel \citep{ftl_sm2025, han2024pfl}. Since both methods adapt the weights of a trained network, they have mostly been implemented with deep learning models, from feed-forward and recurrent networks to Transformers \citep{a1_ann_transfer, a2_transfer_scarcity, a5_federated, zhou2026f2vformer}. On the other hand, \emph{global pooling} trains a single model on the combined records of all vessels and applies it to every vessel without adaptation \citep{nguyen2023multiship}. Differences between vessels are represented by \emph{ship descriptors}, namely a ship identity and ship specifications such as size and propulsion characteristics. Because no adaptation step is required, any regression method can be used, including gradient-boosted trees.

In the literature, multi-ship learning has commonly shown superior performance over per-ship modeling \citep{a1_ann_transfer, a2_transfer_scarcity, federated_green2023, a5_federated, zhou2026f2vformer}, suggesting that knowledge transfer across vessels can lead to better generalization to different operating conditions and even to \emph{unseen vessels}, whose FOC records are not included in training \citep{parkes2022fleet, zhou2026f2vformer}. However, it remains difficult to determine which approach actually performs best, under which conditions, and by how much. No clear and thorough comparison of these approaches is available. The main reason is that studies have used different evaluation protocols, most of which are flawed for pre-departure prediction. First, many studies use inputs that are measured only during the voyage, such as main-engine shaft power, engine RPM, and engine temperature \citep{d5_cnn_lstm, a2_transfer_scarcity, ftl_sm2025}. These variables are almost direct indicators of fuel consumption, and models that use them thus estimate the consumption of a voyage in progress rather than predict it before departure. Second, many studies split the records into training and test sets at random. Because consecutive records of a vessel are nearly identical, a random split places almost the same operating conditions in both sets and thus inflates the reported accuracy relative to chronological validation \citep{vorkapic2026chronological}. Even splits that keep each voyage intact can train on later voyages to predict earlier ones \citep{son2025varnn}, whereas in practice only past voyages are available and the FOC of a vessel changes over time as its hull fouls \citep{schultz2007}. These problems suggest that a fair comparison of FOC prediction approaches is needed under properly controlled experimental conditions.

In this study, we investigate the FOC prediction performance of per-ship, adaptation-based, and global pooling approaches under thoroughly controlled pre-departure conditions and find that global pooling is the best overall approach in every setting examined. To make the comparison fair, all methods use the same fleet of 21 vessels, input variables, data partitions, and metrics. The inputs exclude variables measured only during the voyage, such as engine RPM and shaft power, and each vessel is tested on voyage legs that follow all of its training records. The global model is implemented with five model families, from gradient-boosted trees to a recent Transformer, and compared with per-ship, transfer, and federated baselines, including the vessel-specific adaptation of that Transformer. In particular, we show that the global model performs well in \emph{zero-shot prediction}, where a vessel is predicted without any of its own FOC records, using leave-one-ship-out (LOSO) tests in which each vessel in turn is excluded from training. We further examine how the zero-shot error depends on the specification distance between an unseen vessel and its nearest neighbor in the fleet, whether adaptation helps once the first records of an unseen vessel become available (few-shot prediction), and which features determine accuracy for unseen vessels and for \emph{in-fleet vessels}, whose records are included in training. A plausible explanation for these results is that the physical relationship between speed, loading, weather, and fuel consumption is largely common to all vessels, and the differences between vessels can largely be represented by ship specifications. Vessel-specific adaptation thus adds little information to a global model and may even overfit a vessel with few records.

This study makes three contributions. First, it presents a fleet-wide comparison of per-ship, adaptation-based, and global models under pre-departure conditions. To our knowledge, no such comparison has been reported. Global LightGBM achieves a lower overall mean absolute percentage error (MAPE) than every per-ship, transfer, and federated model, with 8.92\% against 11.14\% for Per-Ship LightGBM. Fine-tuning a global model for each vessel does not improve its accuracy, even with the low-rank adaptation of a recent Transformer. Second, it shows that the advantage of global pooling extends to unseen vessels. In the LOSO tests, 17 of the 21 vessels are predicted with a MAPE below 15\% without any of their own records. Once the first records of an unseen vessel become available, adding them to the pooled training data is as accurate as vessel-specific adaptation and more accurate than per-ship models trained on these records. Third, it identifies the features that determine these results. For in-fleet vessels, ship identity adds little accuracy once ship specifications are included, whereas for unseen vessels only ship specifications support prediction. The zero-shot error is correlated with the specification distance (Spearman $\rho = 0.72$), which can be computed before a vessel enters service. Overall, this study shows that the per-ship models and vessel-specific adaptation commonly adopted in the literature are not necessary for accurate FOC prediction once the approaches are compared under fair pre-departure conditions. A single global model trained on the pooled records of a fleet is sufficient for both in-fleet and newly delivered vessels, which simplifies the development and operation of FOC models for voyage planning and CII compliance. The study also provides an evaluation protocol that excludes information unavailable before departure, which future studies can use to compare FOC prediction approaches on equal terms.

The remainder of this paper is organized as follows. Section~\ref{sec:related} reviews related work. Section~\ref{sec:method} describes the data, the input features, and the compared modeling approaches. Section~\ref{sec:experiments} defines the evaluation protocol and the experiments. Section~\ref{sec:results} reports the results. Section~\ref{sec:conclusion} concludes the paper.

\section{Related Work}
\label{sec:related}

Many approaches to data-driven FOC prediction have been proposed, but their reported accuracies were obtained with different inputs and evaluation protocols. We therefore review the literature with two questions in mind, namely what each modeling approach has shown and under which evaluation conditions it has shown it. We also review the studies that predict vessels without their own records, because such prediction is needed for newly delivered vessels.

\subsection{From Per-Ship to Multi-Ship Modeling}
\label{sec:multiship}
\label{sec:pership_methods}

Per-ship modeling, in which a separate model is trained for each vessel, has been the standard approach to data-driven FOC prediction. The models used in this approach range from support vector regression of shaft power \citep{kim2020svr} to gradient-boosted trees \citep{handayani2025foc, l4_comprehensive_eval, b2_comparison}, deep learning \citep{d5_cnn_lstm, wang2023ga_lstm, b5_transformer_foc}, and physics-informed models of fuel consumption and shaft power \citep{d1_pikan}. The gradient-boosted models are built on algorithms such as XGBoost, LightGBM, and CatBoost \citep{chen2016xgboost, ke2017lightgbm, prokhorenkova2018catboost}. These studies improve the model for a given vessel, but each model still learns only from the records of that vessel.

Multi-ship learning uses the records of other vessels to compensate for the limited records of a single vessel. Its main difficulty is that the vessels differ in size, machinery, and operating profile, so that the knowledge shared across vessels has to account for these differences. The existing studies take two approaches to this difficulty, namely adapting a model learned across vessels to each target vessel and training one global model on the pooled records.

\subsubsection{Adaptation-Based Approaches}

Adaptation-based approaches learn knowledge across vessels and then adapt it to each target vessel with its own records. In transfer learning, a model is first pretrained on data-rich source vessels and then fine-tuned on a data-scarce target vessel, so that the target vessel needs fewer records of its own.

Early studies of transfer learning reported large gains, but their evaluations limit what these gains show. \citet{a1_ann_transfer} transferred a neural network from seven source container vessels to one target vessel with AIS-derived and meteorological inputs without engine RPM. The best of their three transfer strategies reduced the MAPE of the target vessel from 27.03\% for the per-ship model to 11.00\%. However, this comparison used only 100 target records, about 20 of them for testing. The paper also does not report whether its 8:2 hold-out preserved the chronological order. \citet{a2_transfer_scarcity} reduced the MAE by 42\% and 36\% with BiLSTM and random-forest transfer between two vessels, but their inputs include shaft RPM and are therefore not available before departure.

A recent study evaluated transfer learning without post-event inputs and with temporal separation, but only for target vessels excluded from pretraining. \citet{zhou2026f2vformer} pretrained a Transformer on noon reports from the source vessels of a 170-vessel container fleet, conditioned it on 15 static attributes such as deadweight, engine rating, and service route, and adapted it to each target vessel with low-rank adaptation. The zero-shot MAPE of 13.82\%, averaged over the target vessels, fell to 5.64\% after adaptation with 500 target records. A pooled XGBoost baseline with concatenated static attributes, in contrast, improved only from 14.11\% to 12.15\% as target records were added. Adaptation therefore helped vessels whose records were absent from pretraining. However, the study included no federated alternatives and trained its per-ship baselines on at most 500 target records.

Federated learning addresses a different situation, in which operators cannot share their records. In this setting, the records remain with each operator and only model updates are shared \citep{mcmahan2017fedavg}. The maritime studies of federated learning report gains over individual training, but their evaluations remain limited. \citet{federated_green2023} combined federated prediction with speed optimization for 20 simulated vessel clients. Federated learning outperformed individual training in their study, but the data were entirely synthetic and split at random into 60/20/20 parts. \citet{a5_federated} proposed federated learning with adaptive regularization and CNN-GRU prediction for eight bulk carriers and reduced the RMSE by 0.59--5.93\% relative to individual, FedAvg, and standard-regularization baselines. The temporal separation of their test records was not reported.

Federated transfer and personalized federated learning add vessel-specific adaptation to this setting. \citet{ftl_sm2025} combined random-forest ship mapping with a domain-knowledge ANN under federated transfer learning for 40 vessels of four types and achieved an average MAPE of 8.53\%, but their inputs include the daily mean engine RPM. \citet{han2025marl_twin} personalized federated predictors of ship fuel consumption, trained on seven LNG carriers, to two target LNG carriers and used them for speed control with multi-agent reinforcement learning. Outside the maritime domain, cross-building transfer and personalized federated learning have likewise outperformed local models under data-scarce conditions \citep{k1_cross_building, k2_federated_building}.

Differences between vessels limit the effectiveness of parameter averaging and thus require personalization that makes federated frameworks more complex. Where the records of a fleet can be combined in one place, these differences raise the question of whether a global model can represent them without a separate adaptation step.

\subsubsection{Global Pooling}

Global pooling trains one model on the combined records of several vessels and applies it to every vessel without a vessel-specific adaptation step. The differences between vessels are represented in the inputs, for example by ship descriptors, or in the architecture of the model.

Previous studies of global pooling have shown that one model can be trained on the records of a fleet, but they have not compared it with per-ship or adaptation-based models. \citet{nguyen2023multiship} trained XGBoost and feed-forward neural networks on the noon reports of 56 container vessels, with GT and propeller pitch as ship descriptors together with operational and environmental inputs. Their evaluation tested the robustness of the global models to missing inputs and noise. \citet{noise_robustness2025} similarly trained domain-knowledge ANN and BiLSTM models on the records of 32 container vessels, with GT representing vessel size, and found the vessel speed to be the most sensitive input. \citet{ghosh2026unified} compared 11 learning algorithms for a unified model of a heterogeneous fleet, with inputs that include direct engine load indicators. Their SHAP analysis identified vessel geometry, such as deadweight and breadth, as the main driver of fuel consumption across vessels. These studies thus evaluate the global model only within the pooled setting.

Global pooling has also served as a baseline in a transfer study, in which it was less accurate than transfer but had no ship descriptors. \citet{a2_transfer_scarcity} mixed 3{,}005 records of one container vessel with 489 records of another and trained a single random forest and a single BiLSTM on the mixture. Under five-fold expanding-window evaluation, these models were less accurate than both the transfer models and the models trained on the target vessel alone. However, the mixed models received no ship identity or specifications to distinguish the two vessels. Their eight inputs also included shaft RPM. The result therefore shows that pooling two vessels without ship descriptors can fail, but it does not test pooling with ship descriptors across a fleet under pre-departure conditions.

\subsection{Evaluation Protocols and Data Leakage}
\label{sec:eval_lit}

The reported accuracy of FOC models depends strongly on how the models are evaluated. In particular, leakage of information through the inputs or through the time order of the records inflates the reported accuracy and makes it difficult to judge the value of a method for voyage planning.

Input leakage occurs when a model uses variables that are measured only during the voyage. Several studies include engine telemetry measured at the same time as the FOC, such as engine RPM, engine load, or shaft power \citep[e.g.,][]{mohamed2026datafusion}. These variables are strongly associated with FOC, but their measured values are unavailable when a voyage is planned, so results that use them do not represent the accuracy before departure. Table~\ref{tab:lit_comparison} shows that seven of the 19 reviewed studies use such variables or the measured FOC of earlier time steps as inputs.

Temporal leakage occurs when the test records are not separated in time from the training records. Ship performance changes over time because hull fouling increases the resistance between maintenance events, seasonal weather alters the operating conditions, and operational policies shift the speed distribution. For example, \citet{schultz2007} estimated that light slime increases the total resistance of an FFG-7 frigate at 15 knots by about 11\%. Studies of concept drift \citep{f1_concept_drift, f2_proactive_drift} likewise show that the distribution of operational data shifts over time. Random splits of the records therefore place temporally adjacent records and similar operating conditions in both the training and test sets and thus overestimate the accuracy on future voyages.

Studies of leakage confirm this overestimation. In the broader machine-learning literature, leakage systematically inflates the reported prediction performance \citep{rosenblatt2024leakage}. \citet{albelali2025hidden_leaks} reported RMSE-gain differences of up to 20.5\% between leaky and leakage-free sequence construction under 10-fold cross-validation. For maritime data, \citet{vorkapic2026chronological} found that random partitioning inflated $R^2$ by up to 0.19 relative to chronological validation. Nevertheless, many FOC studies still use random or non-temporal $k$-fold splits, although some, such as \citet{short_term_foc2024}, use a chronological 70/30 hold-out by voyage.

Few studies avoid both kinds of leakage. Table~\ref{tab:lit_comparison} shows that only \citet{b1_fuelcast}, \citet{d1_pikan}, and \citet{zhou2026f2vformer} restrict their inputs to variables available before departure and also separate their test records in time. The FuelCast benchmark of \citet{b1_fuelcast} uses five-fold blocked temporal cross-validation on 173{,}986 observations from three vessels but models each vessel independently. In four of its five folds, however, the training intervals include records later than the test interval. \citet{d1_pikan} likewise model each of five cargo vessels separately and test each vessel on its later records. \citet{son2025varnn} also restricted the inputs and proposed a voyage-aware recurrent model (VaRNN) for the distribution shift between voyages, but its five folds were drawn at random from 16 voyages of one bulk carrier and thus did not separate past and future. Apart from \citet{zhou2026f2vformer}, no study of multi-ship learning has therefore avoided both kinds of leakage.

\subsection{Zero-Shot Prediction for Unseen Vessels}
\label{sec:zeroshot_lit}

A newly delivered vessel has no records of its own, so its FOC can only be predicted by a model trained on other vessels. In zero-shot prediction, all FOC records of the target vessel are excluded from training and validation, so that the vessel is predicted only from its specifications and operating inputs without fitting any vessel-specific parameters.

Previous studies have shown that zero-shot prediction is possible. \citet{nguyen2023multiship} excluded each of eight ship-size groups in turn from their pooled XGBoost and ANN models and predicted the excluded group without fine-tuning, with GT and propeller pitch describing the unseen vessels. \citet{parkes2022fleet} predicted the shaft power of each of nine LNG carriers with neural networks trained on the other eight, using five inputs without ship descriptors. The errors were about 4\% for six of the nine vessels. The larger errors of up to 9.3\% were attributed to a different propulsion system and to operating conditions outside the training data. Because the vessels were of similar size, the authors concluded that coverage of the operating conditions mattered more than vessel parameters for that fleet. Whether this conclusion holds for a fleet spanning a wide range of vessel sizes remains open. \citet{d4_hybrid_physics} took a different route with a fleet-oriented hybrid framework, in which random-forest components estimate the resistance and propeller diameter within a physical FOC calculation. Its validation, however, covered FOC for two container vessels and power for two bulk carriers. \citet{zhou2026f2vformer} reported zero-shot MAPEs between 13.45\% and 14.36\% across the four capacity classes of their fleet.

These studies, however, do not establish which vessel information supports zero-shot prediction or how its accuracy varies among individual vessels. \citet{nguyen2023multiship} assessed the dependence on specifications in feature-removal tests that were separate from the size-group holdouts, so their results cannot show whether specifications become more important when the records of the target vessel are excluded. \citet{zhou2026f2vformer} found that removing the static attributes raised the zero-shot MAPE from 13.82\% to 15.23\%, but their target vessels were always excluded from pretraining, so this contribution was not compared with in-fleet prediction. In their size-group experiments, \citet{nguyen2023multiship} also found that removing one source group improved ANN prediction for another excluded group. This finding links the accuracy to the composition of the training fleet. However, both studies reported errors only by group, so neither related the error of an individual vessel to its similarity to the training fleet. An analysis of individual vessels across a fleet with widely different specifications is therefore needed.

\subsection{Summary and Research Gap}
\label{sec:gap}

Table~\ref{tab:lit_comparison} summarizes the reviewed studies by modeling approach, data scale, inputs, and evaluation protocol.

\begin{table}[htbp]
\centering
\caption{Methodological comparison of FOC prediction studies. $\checkmark$/$\times$: criterion met/not met; --: not reported or not applicable.}
\label{tab:lit_comparison}
\footnotesize
\renewcommand{\arraystretch}{1.15}
\setlength{\tabcolsep}{1.5pt}
\resizebox{\textwidth}{!}{%
\begin{tabular}{llr r l rr cc l}
\toprule
Study & Approach & $N_\text{v}$ & Period & Interval (R/P) & $N_\text{raw}$ & $N_\text{obs}$ & Plan.\textsuperscript{$\dagger$} & Temp.\textsuperscript{$\ddagger$} & Evaluation \\
\midrule
\citet{b2_comparison} & Per-ship & 1 & $\sim$18\,mo. & daily / daily & 335 & 168 & $\checkmark$ & -- & 8:2 hold-out \\
\citet{d5_cnn_lstm} & Per-ship & 1 & 29\,mo. & 60\,s / -- & 1.0M & -- & $\times$ & $\checkmark$ & 5-fold CV + temporal holdout \\
\citet{wang2023ga_lstm} & Per-ship & 1 & $\sim$7.5\,mo. & 10\,min / 1\,h & -- & 3{,}361 & $\checkmark$ & $\times$ & interleaved split \\
\citet{short_term_foc2024} & Per-ship & 1 & 14--17\,mo. & 30\,s / 1\,min & 1{,}237{,}562 & 65{,}560 & $\times$ & $\checkmark$ & 70/30 voyage hold-out \\
\citet{son2025varnn} & Per-ship & 1 & 18\,mo. & 10\,s / 10\,s & -- & 2.4M & $\checkmark$ & $\times$ & 5-fold voyage CV \\
\citet{mohamed2026datafusion} & Per-ship & 1 & 12\,mo. & daily / daily & 296 & 266 & $\times$ & $\checkmark$ & 80:20 random split + chronological split \\
\citet{b1_fuelcast} & Per-ship & 3 & 3--12\,mo. & 5\,min / 5\,min & 173{,}986 & 173{,}986 & $\checkmark$ & $\checkmark$ & 5-fold temporal CV \\
\citet{d1_pikan} & Per-ship & 5 & $\sim$17--61\,mo. & 15\,min / 15\,min & -- & 386{,}259 & $\checkmark$ & $\checkmark$ & chronological split per vessel \\
\midrule
\citet{a1_ann_transfer} & Transfer & 8 & $\sim$21\,mo. & -- / daily & 801{,}665 & 4{,}710 & $\checkmark$ & -- & 8:2 hold-out \\
\citet{a2_transfer_scarcity} & Transfer & 2 & 8\,d--5\,mo. & -- / -- & -- & 3{,}494 & $\times$ & $\checkmark$ & 5-fold expanding-window CV \\
\citet{zhou2026f2vformer} & Transfer & 170 & -- & daily / daily & -- & -- & $\checkmark$ & $\checkmark$ & vessel hold-out + last 200 records \\
\citet{federated_green2023} & Federated & 20 & 6\,mo. & -- / daily & -- & 3{,}600 & $\times$ & $\times$ & 60/20/20 random split \\
\citet{a5_federated} & Federated & 8 & 12\,mo. & -- / 6\,h & -- & 7{,}107 & $\checkmark$ & -- & -- \\
\citet{ftl_sm2025} & Federated transfer & 40 & 39\,mo. & -- / daily & -- & -- & $\times$ & -- & hold-out \\
\citet{han2025marl_twin} & Federated & 7 & 12\,mo. & -- / 6\,h & -- & -- & $\checkmark$ & -- & 80/10/10 hold-out \\
\citet{ghosh2026unified} & Global & 3 & 24\,mo. & 4--25\,h / segment & -- & $\sim$1{,}100 & $\times$ & $\times$ & 80:20 voyage-group split (5-fold CV) \\
\citet{d4_hybrid_physics} & Global hybrid & 4 & 16--48\,mo. & -- / -- & -- & 866 & $\checkmark$ & -- & case-study validation \\
\citet{noise_robustness2025} & Global & 32 & 18\,mo. & daily / -- & 11{,}155 & -- & $\checkmark$ & -- & 80:20 split (5-fold CV) \\
\citet{nguyen2023multiship} & Global & 56 & 30\,mo. & daily / daily & 87{,}894 & 21{,}415 & $\checkmark$ & -- & 75/25 + size-group holdout \\
\midrule
\textbf{This study} & \textbf{Global} & \textbf{21} & \textbf{55\,mo.} & \textbf{10\,min / 3\,h} & $\sim$\textbf{4.0M} & \textbf{130K} & $\checkmark$ & $\checkmark$ & \textbf{temp.\ holdout} \\
\bottomrule
\end{tabular}}
\\[4pt]
{\footnotesize
\textsuperscript{$\dagger$}Inputs available before departure (no realized engine RPM or power, no measured FOC). \;
\textsuperscript{$\ddagger$}Temporal train--test separation (no random shuffling). \\
Interval (R/P): R: raw logging rate of the original records; P: rate after preprocessing. $N_\text{raw}$: records as originally logged; $N_\text{obs}$: observations used for modeling after preprocessing.
}
\end{table}

Because these studies differ in features, evaluation protocols, and fleets, their reported accuracies cannot be compared directly. As a result, it remains unclear whether global pooling is more accurate than per-ship and adaptation-based approaches when all of them are evaluated under the same pre-departure conditions. The pooled studies show that a global model can be trained on fleet records but do not compare it with the alternatives on future voyages. \citet{a2_transfer_scarcity} compared pooling and adaptation directly, but their pooled baseline had no ship descriptors, covered two vessels, and used shaft RPM. \citet{zhou2026f2vformer} made a broader comparison without post-event inputs, but for data-scarce target vessels excluded from pretraining, without federated alternatives, and with per-ship baselines limited to 500 records. The open question is whether pooling improves prediction across a fleet whose vessels all have operating histories and whether vessel-specific adaptation adds accuracy once the records of a vessel are pooled. We therefore compare per-ship, transfer, federated, and global models, including a re-implementation of the Transformer of \citet{zhou2026f2vformer}, on a common fleet with the same inputs, voyage-leg partitions, prediction target, and evaluation metrics.

For zero-shot prediction, the remaining questions are which vessel information supports it and how its accuracy varies among individual unseen vessels. The specification-based global models and size-group tests of \citet{nguyen2023multiship} provide a basis for these questions. We therefore examine how the contributions of ship identity and ship specifications differ between in-fleet and zero-shot prediction and relate the zero-shot error of each vessel to its specification distance from the rest of the fleet.

\section{Methodology}
\label{sec:method}

This section describes the dataset, the input features, and the modeling approaches compared in this study. Section~\ref{sec:dataset} introduces the 21-vessel dataset, and Section~\ref{sec:preprocessing} describes its preprocessing. Section~\ref{sec:features} describes the input features available before departure, including the ship descriptors that distinguish the vessels, and Section~\ref{sec:global_framework} formulates per-ship modeling, adaptation-based approaches, and global pooling in a common notation.

\subsection{Dataset}
\label{sec:dataset}

The data were collected from 21 vessels of a single operator, 20 bulk carriers and one oil tanker, between January 2021 and July 2025 (55 months). Table~\ref{tab:vessel_classes} shows the composition of the fleet by vessel class, with deadweight ranging from about 110{,}000 to 400{,}000\,DWT (3.7$\times$) and main-engine power from about 13{,}000 to 29{,}000\,kW (2.2$\times$). The individual ship specifications are used as model inputs but cannot be disclosed under a confidentiality agreement with the operator, and Table~\ref{tab:vessel_classes} thus reports rounded class-level values. The fleet operates mainly between Asia and South America, although some vessels also sail to Europe, the Middle East, and Oceania.

\begin{table}[htbp]
\centering
\caption{Fleet composition by vessel class. Specifications are approximate values representative of each class. All values are rounded to preserve confidentiality.}
\label{tab:vessel_classes}
\footnotesize
\begin{tabular}{llrrrrrr}
\toprule
Class & Type & $N_\text{v}$ & DWT & MCR & LBP & $B$ & Age \\
 & & & ($\sim$kDWT) & ($\sim$kW) & ($\sim$m) & ($\sim$m) & (yr) \\
\midrule
A & Bulk carrier & 8  & 400 & 29{,}000 & 350 & 65 & 9--15 \\
B & Bulk carrier & 11 & 320 & 21{,}000 & 330 & 62 & 4--7 \\
C & Bulk carrier & 1  & 180 & 19{,}000 & 280 & 45 & $\sim$15 \\
D & Oil tanker   & 1  & 110 & 13{,}000 & 240 & 44 & $\sim$8 \\
\midrule
 & \textbf{Total} & \textbf{21} & & & & & 4--15 \\
\bottomrule
\end{tabular}
\end{table}

Onboard sensors logged about 4.0 million records at 10-minute intervals over this period. Each record contains the main-engine FOC in tonnes per hour together with navigational and operational variables. The environmental conditions, namely wind, current, waves, and sea surface temperature, were obtained from gridded meteorological forecast products and interpolated to the position and time of each record.

\subsection{Data Preprocessing}
\label{sec:preprocessing}

\begin{figure}[htbp]
    \centering
    \includegraphics[width=\textwidth]{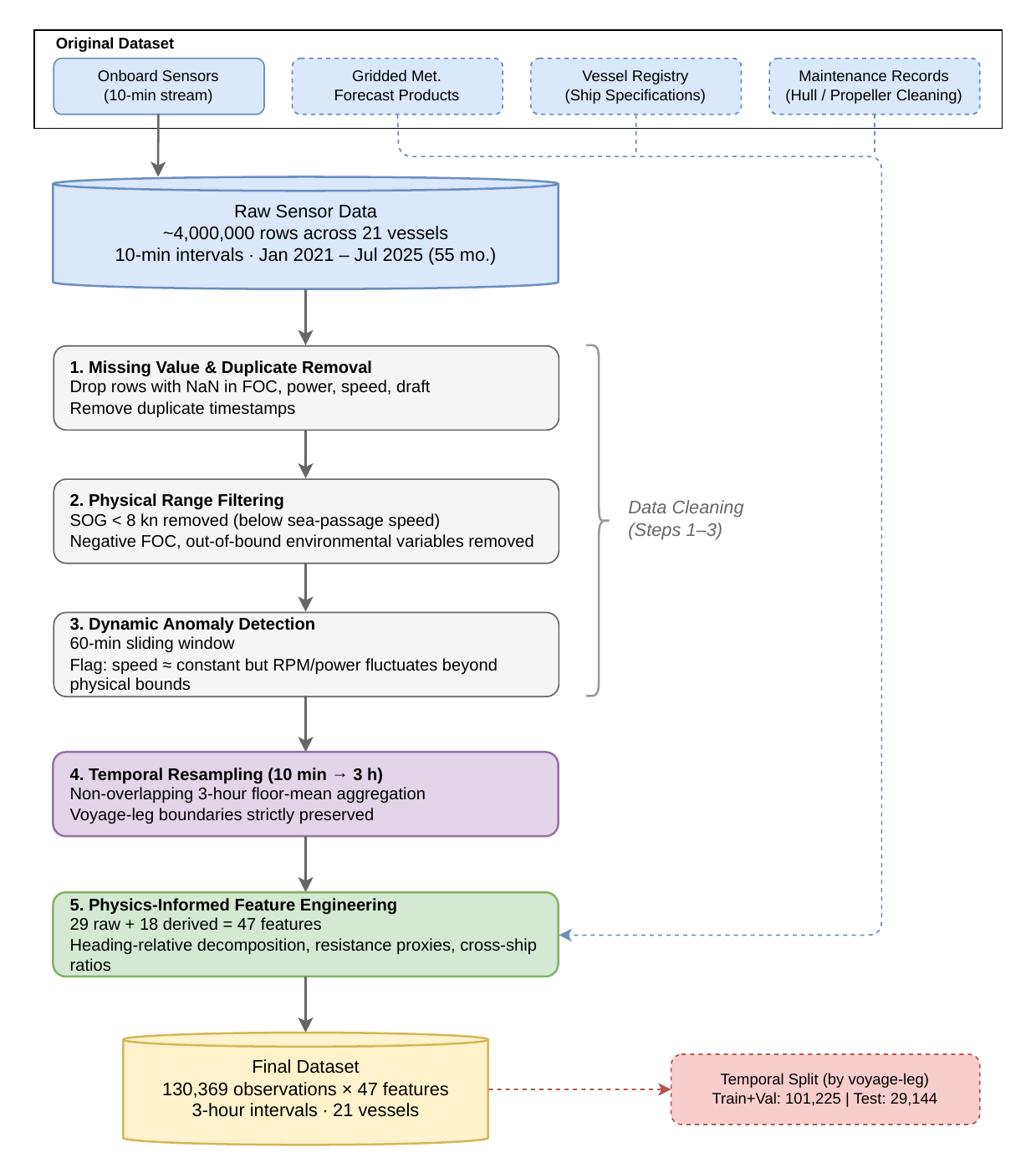}
    \caption{Data preprocessing pipeline. The raw 10-minute sensor records (about 4.0 million) are cleaned in three stages, averaged over 3-hour windows within voyage legs, and extended with derived features, resulting in 130{,}369 observations with 47 features.}
    \label{fig:preprocessing}
\end{figure}

Fig.~\ref{fig:preprocessing} shows the preprocessing pipeline, in which the raw records were cleaned in three stages before modeling. First, records with missing FOC, engine power, speed, or draft and records with duplicate timestamps were removed. Next, physical range limits were applied to remove negative FOC, environmental values outside physical bounds, and speeds below 8 knots, a threshold set with ship-domain experts to retain sea passages and exclude port, maneuvering, and drifting periods. Finally, records showing sensor inconsistencies within 60-minute windows were removed, for example when engine RPM or power fluctuates while the speed remains nearly constant. Engine RPM and power were used only for this cleaning and are not model inputs. The cleaned records were then averaged over non-overlapping 3-hour windows that do not cross voyage-leg boundaries. This averaging reduces short-term fluctuations in the 10-minute sensor data and matches the time scale of voyage planning, in which speed and weather are specified over hours rather than minutes. The derived features of Section~\ref{sec:features} were computed from the averaged records.

The resulting dataset contains 130{,}369 observations at 3-hour intervals, which are used in all experiments. The FOC in this dataset ranges from 0.28 to 4.25\,t/h, with a fleet mean of 2.02\,t/h. Table~\ref{tab:ship_data} shows the number of observations per vessel, which ranges from 490 (Ship~8) to 9{,}935 (Ship~18) because the recording periods differ.

\begin{table}[htbp]
\centering
\caption{Number of observations per vessel. $n_{\text{train}}$: observations in the training and validation partitions; $n_{\text{test}}$: observations in the temporal test set (Section~\ref{sec:eval_protocol}). Vessels are anonymized as Ship~1--21.}
\label{tab:ship_data}
\footnotesize
\begin{tabular}{lcrrclcrr}
\toprule
Vessel & Class & $n_{\text{train}}$ & $n_{\text{test}}$ & \hspace{6pt} & Vessel & Class & $n_{\text{train}}$ & $n_{\text{test}}$ \\
\midrule
Ship 1  & B & 4{,}304  & 1{,}304 & & Ship 12 & B & 5{,}715  & 1{,}842 \\
Ship 2  & B & 7{,}392  & 1{,}917 & & Ship 13 & A & 600     & 224 \\
Ship 3  & D & 5{,}877  & 1{,}530 & & Ship 14 & A & 5{,}983  & 1{,}878 \\
Ship 4  & B & 4{,}826  & 1{,}523 & & Ship 15 & A & 1{,}774  & 501 \\
Ship 5  & B & 5{,}918  & 1{,}631 & & Ship 16 & A & 2{,}877  & 856 \\
Ship 6  & B & 7{,}717  & 2{,}134 & & Ship 17 & B & 5{,}776  & 1{,}713 \\
Ship 7  & C & 6{,}680  & 1{,}970 & & Ship 18 & B & 7{,}866  & 2{,}069 \\
Ship 8  & A & 247     & 243    & & Ship 19 & A & 3{,}211  & 830 \\
Ship 9  & B & 4{,}044  & 1{,}143 & & Ship 20 & B & 6{,}887  & 2{,}089 \\
Ship 10 & A & 3{,}861  & 1{,}009 & & Ship 21 & B & 3{,}197  & 924 \\
Ship 11 & A & 6{,}473  & 1{,}814 & & \textbf{Total} & & \textbf{101{,}225} & \textbf{29{,}144} \\
\bottomrule
\end{tabular}
\end{table}

\subsection{Feature Design}
\label{sec:features}

Table~\ref{tab:features} shows the 47 input features used in this study. The ship descriptors identify each vessel by its ship identity and seven ship specifications, and the operating and environmental variables describe each 3-hour observation by the speed, course, draft, cargo, weather and ocean conditions, water depth, and hull and propeller cleaning history. The within-observation physics features and the cross-ship ratios are derived from these variables. These features were designed as follows.

First, we chose the input variables so that all of them are available when a voyage is planned. For the operating conditions, we chose the speed, course, draft, and cargo quantity because they are fixed in the voyage and loading plans. For the environmental conditions, three sources are possible, namely onboard measurements, reanalysis (hindcast) data, and weather and ocean forecasts. Onboard measurements and reanalysis data describe the conditions actually encountered, but they become available only during or after the voyage. We therefore chose forecast products, which provide the wind, current, waves, and sea surface temperature along the planned route before departure. We also excluded main-engine RPM and shaft power because, although closely related to FOC, they are known only during the voyage. Accordingly, the derived features were computed only from these chosen inputs.

Second, we described each vessel by two kinds of ship descriptors so that a single model can distinguish vessels whose fuel demand differs under the same operating conditions. The ship identity is a categorical feature that allows the model to learn the individual characteristics of each vessel in the training fleet. However, it carries no information about a vessel that is not included in training. We therefore also chose seven ship specifications, such as the maximum continuous rating (MCR), deadweight (DWT), and propeller dimensions, because they describe every vessel on a common physical scale and are available even for an unseen vessel. In addition, we added five cross-ship ratios, such as the speed relative to the square root of the length and the draft relative to the DWT, which express the operating conditions relative to the size of the vessel.

Third, we represented the environmental conditions relative to the vessel. The course and the wind, current, and wave directions were first encoded by their sine and cosine because their raw angles are discontinuous at 0$^\circ$ and 360$^\circ$. Since the effect of wind, waves, and currents on resistance depends on their direction relative to the heading, they were then decomposed into components along and across the heading. In addition, the speed through water was estimated by subtracting the along-track current from the speed over ground. Together with the powers of speed $V^2$ and $V^3$, which reflect the nonlinear increase of resistance with speed \citep{holtrop1982}, and four proxies for environmental loading, these terms form the 13 within-observation physics features.

Finally, we included the days since the last hull and propeller cleaning because fouling increases the resistance, and thus the FOC, of a vessel under the same operating conditions as time passes \citep{schultz2007}. Flags also indicate whether cleaning records exist. For the six vessels without maintenance records, the flags were set to zero and the days since cleaning to the fleet median.

\begin{table}[htbp]
\centering
\caption{The 47 features by group. For raw inputs, the last column gives where the value comes from before departure; derived features are computed from the raw inputs. Main-engine RPM and shaft power are excluded. Ship age is computed at 1 March 2026. The groups are those analyzed in the ablation (Section~\ref{sec:ablation}).}
\label{tab:features}
\footnotesize
\begin{tabular}{@{}lrp{4.6cm}p{5.9cm}@{}}
\toprule
Group & $n$ & Features & Source before departure / description \\
\midrule
\multicolumn{4}{@{}l}{\textbf{Ship descriptors (8)}} \\
\quad Ship identity & 1 & Ship name & Vessel registry \\
\quad Ship specifications & 7 & MCR, DWT, LBP, breadth, propeller diameter, propeller pitch, ship age & Vessel registry and design documents \\
\midrule
\multicolumn{4}{@{}l}{\textbf{Operating state \& environment (21)}} \\
\quad Operational & 5 & SOG, course (sin/cos), mean draft, cargo quantity & Voyage plan (speed, course); loading plan (draft, cargo) \\
\quad Environmental & 12 & Wind speed \& dir.\ (sin/cos), current speed \& dir.\ (sin/cos), wave height, dir.\ (sin/cos) \& period, water depth, SST & Weather and ocean forecasts along the planned route; charts (water depth) \\
\quad Maintenance & 4 & Hull cleaning (flag, days since), propeller cleaning (flag, days since) & Maintenance records \\
\midrule
\multicolumn{4}{@{}l}{\textbf{Within-observation physics (13)}} \\
\quad Heading-relative & 7
  & headwind, crosswind & wind along/across heading \\
 && head wave, beam wave & wave along/across heading \\
 && along-track, cross-track current & current along/across heading \\
 && STW & estimated speed through water \\
\addlinespace[2pt]
\quad Resistance \& interaction & 6
  & $V^2$, $V^3$ & speed--resistance power-law terms \\
 && $H_s^2/T_p$ & wave-added resistance proxy \\
 && $W_{\text{head}} \!\times\! C_{\text{along}}$ & current--wind interaction (combined head forces) \\
 && $H_s / V$ & wave--speed ratio (wave severity relative to vessel speed) \\
 && $\sqrt{H_s^2 + (U_w/10)^2}$ & sea state index (combined wave height and wind speed) \\
\midrule
\multicolumn{4}{@{}l}{\textbf{Cross-ship ratios (5)}} \\
\quad Scale-related combinations & 5
  & $V^3\!/P_{\text{MCR}}$ & speed--power ratio \\
 && $T/\mathrm{DWT}$ & loading state \\
 && $V/\!\sqrt{L_{\text{BP}}}$ & Froude number proxy \\
 && $V^2 \!\times\! T$ & frictional resistance $\times$ loading \\
 && $V_{\text{STW}}/V_{\text{SOG}}$ & current effect (size-independent) \\
\midrule
\textbf{Total} & \textbf{47} & & \\
\bottomrule
\end{tabular}
\\[4pt]
{\footnotesize
Symbols: $V$: speed over ground (SOG); $H_s$: significant wave height; $T_p$: wave period; $U_w$: wind speed;
$W_{\text{head}}$: headwind component; $C_{\text{along}}$: along-track current; $T$: mean draft;
DWT: deadweight tonnage; $P_{\text{MCR}}$: maximum continuous rated power;
$L_{\text{BP}}$: length between perpendiculars; $V_{\text{STW}}$: speed through water.}
\end{table}

\subsection{Modeling Approaches}
\label{sec:global_framework}

\begin{figure}[!t]
    \centering
    \includegraphics[width=\textwidth]{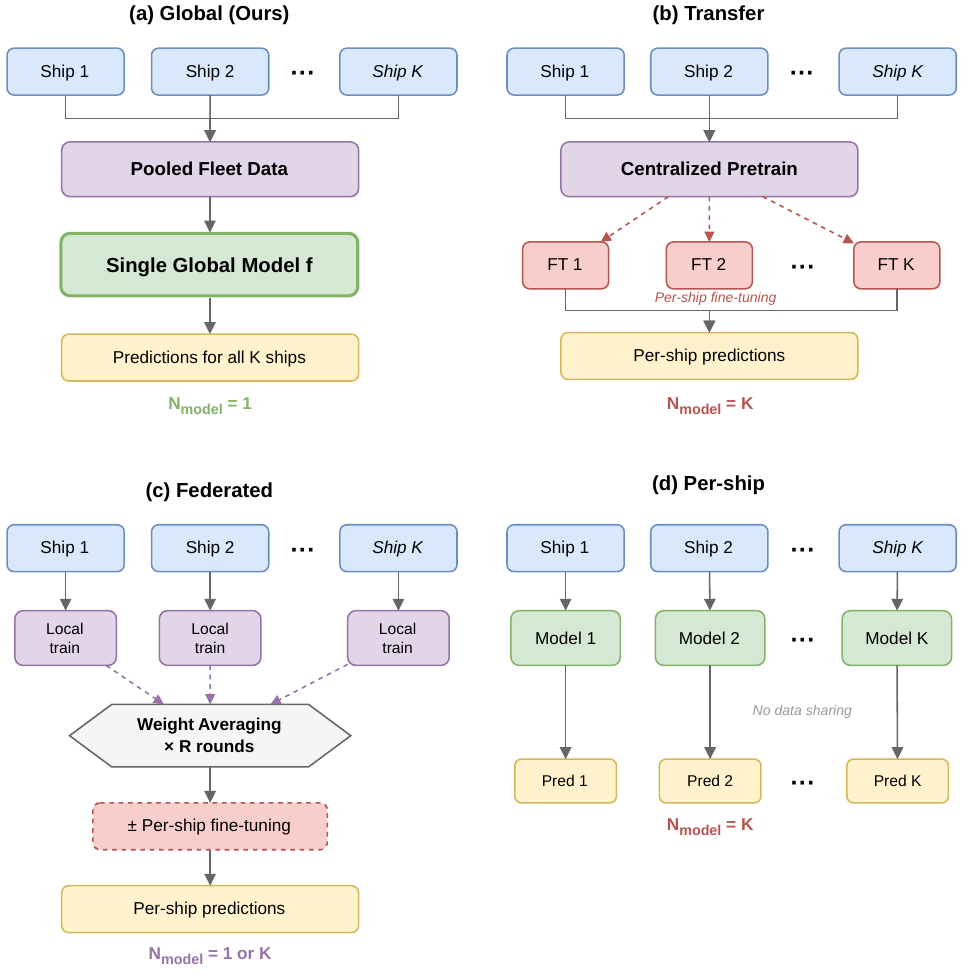}
    \caption{Training procedures of the compared methods: global pooling (a), the adaptation-based approaches of transfer (b) and federated learning (c), and per-ship modeling (d). (a)~Global: a single model is trained on pooled fleet data. (b)~Transfer: centralized pretraining followed by per-ship fine-tuning. (c)~Federated: local training with weight averaging across $R$ communication rounds, optionally followed by per-ship fine-tuning. (d)~Per-ship: independent models trained on each vessel's records with no data sharing.}
    \label{fig:paradigm_comparison}
\end{figure}

\begin{figure}[!t]
    \centering
    \includegraphics[width=\textwidth]{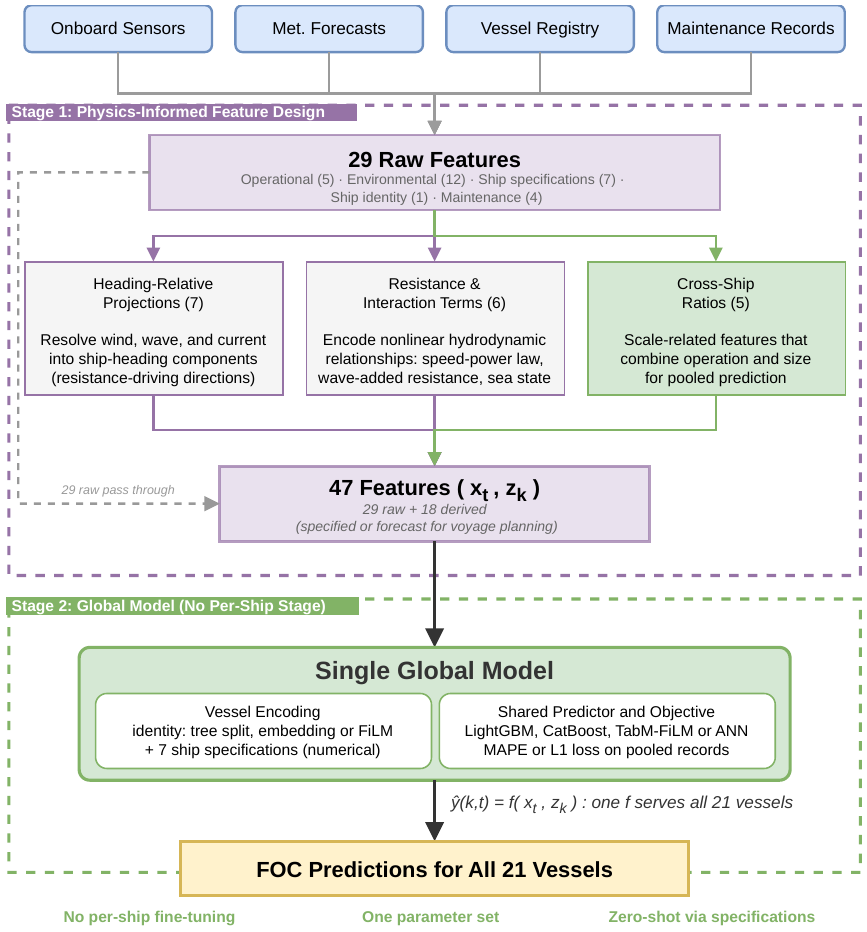}
    \caption{Architecture of the global model. Four data sources feed a feature design stage that produces the 47 features of Table~\ref{tab:features}, namely 29 raw inputs (the ship descriptors and the operating and environmental variables) and 18 derived features (the within-observation physics features and the cross-ship ratios). One model, encoding each vessel by its ship identity and ship specifications, is trained on the pooled records of all 21 vessels and predicts FOC without a vessel-specific stage. For an unseen vessel, the identity is unknown and the prediction rests on its ship specifications and operating inputs (zero-shot prediction).}
    \label{fig:method_overview}
\end{figure}

The three modeling approaches compared in this study use the same input features but differ in how they combine the records of different vessels and where they keep vessel-specific information. Fig.~\ref{fig:paradigm_comparison} shows their training procedures, which we formulate below in a common notation. For each vessel $k=1,\dots,K$, let $D_k$ denote its records and $\mathbf{x}_t$ the operating, environmental, and derived features of Section~\ref{sec:features} at time $t$.

In per-ship modeling, shown in Fig.~\ref{fig:paradigm_comparison}d, a separate model $f_k$ is trained on the records $D_k$ of each vessel and predicts $\hat{y}_{k,t}=f_k(\mathbf{x}_t)$. Vessel-specific information is thus kept in $K$ separate models, and no information is shared between vessels. A per-ship model, however, cannot be built for a vessel that has no records.

In adaptation-based approaches, a model is first learned from several vessels and then adapted to each vessel with its own records, that is, $\hat{y}_{k,t}=f(\mathbf{x}_t;\theta_k)$, where the vessel-specific parameters $\theta_k$ are obtained by adjusting the shared parameters $\theta$ with $D_k$. In transfer learning, shown in Fig.~\ref{fig:paradigm_comparison}b, $\theta$ is obtained by pretraining on the records of several vessels and $\theta_k$ by fine-tuning on $D_k$. In federated learning, shown in Fig.~\ref{fig:paradigm_comparison}c, $\theta$ is instead obtained by averaging the parameters of local models trained on each vessel without collecting their records, and $\theta_k$ by fine-tuning or personalization. Vessel-specific information is thus kept in the adapted parameters $\theta_k$.

In global pooling, shown in Fig.~\ref{fig:paradigm_comparison}a, a single model $f$ is trained on the combined records of all vessels and predicts $\hat{y}_{k,t}=f(\mathbf{x}_t,\mathbf{z}_k)$, where $\mathbf{z}_k$ denotes the ship descriptors of vessel $k$. Vessel-specific information is thus supplied as an input rather than kept in separate or adapted parameters, and the same parameters are used for every vessel without an adaptation step. Fig.~\ref{fig:method_overview} shows the architecture of the global model used in this study. Since the ship specifications in $\mathbf{z}_k$ are known before a vessel has any FOC records, the global model can also predict a vessel without any of its records. Per-ship modeling cannot, and adaptation-based approaches need the records of the vessel for their adaptation step.

\FloatBarrier
\section{Experiments}
\label{sec:experiments}

This section describes how the comparison was carried out. Section~\ref{sec:eval_protocol} defines the data partition, Section~\ref{sec:metrics} the evaluation metrics, and Section~\ref{sec:models} the implementation of the compared methods. Sections~\ref{sec:infleet}--\ref{sec:retraining_method} then describe the analyses of the in-fleet comparison, zero-shot and few-shot prediction, the factors that contribute to the accuracy of the global model, and quarterly retraining.

\subsection{Data Partition}
\label{sec:eval_protocol}

\begin{figure}[!t]
    \centering
    \includegraphics[width=0.92\textwidth]{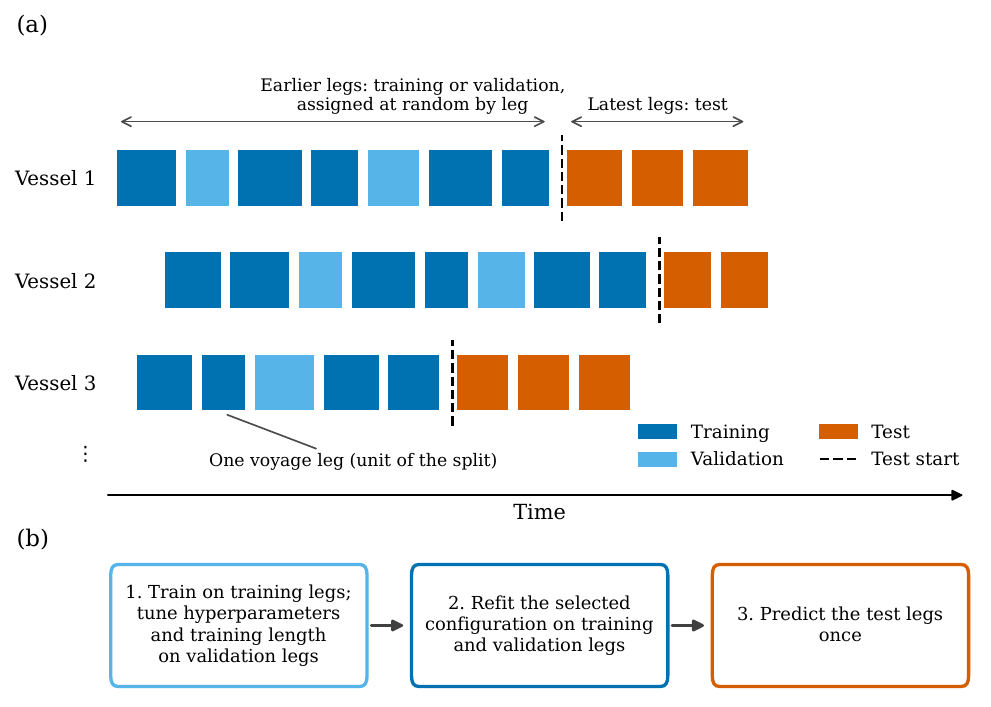}
    \caption{Data partition. (a)~The records of each vessel are split by voyage leg. The latest legs of each vessel form the test set, and the earlier legs are assigned at random, by whole leg, to training or validation. (b)~Training procedure of the main comparison.}
    \label{fig:data_partition}
\end{figure}

The way the records are divided into training and test data determines whether the evaluation reflects prediction before departure. If a model is tested on periods it has already seen during training, its reported accuracy overstates what is achievable when future voyages are planned. We therefore divided the records of each vessel so that it is tested on voyages that follow all of its training records, as shown in Fig.~\ref{fig:data_partition}a. The voyage leg, a continuous sea passage between two port calls (median 105 observations, about 13 days), was used as the unit of the split, because consecutive records within a leg are strongly correlated and would otherwise fall into both training and test data. The latest legs of each vessel, defined by the data provider, form the test set, and the earlier legs were assigned at random, by whole leg and with a fixed seed, to the training and validation sets. Table~\ref{tab:partition} gives the number of legs and observations in each partition.

In the main comparison, all methods used this partition and predicted the test legs only once, following the procedure in Fig.~\ref{fig:data_partition}b. The models were first trained on the training legs, and the validation legs were used to select their training length and, where tuned, their hyperparameters. The selected configuration was then refitted on the training and validation legs together, and the refitted model predicted the test legs. The federated and fine-tuned baselines follow this protocol with small differences in the final fitting step, which are described in Appendix~\ref{app:implementation}.

Because the test periods of the vessels begin at different dates, the training data of a global model can include records of other vessels dated within the test period of a target vessel. To confirm that these records do not explain the results, we also refitted Global LightGBM for each target vessel without them.

\begin{table}[htbp]
\centering
\caption{Data partitions. Voyage legs are never split between partitions; the test legs are the latest legs of each vessel.}
\label{tab:partition}
\footnotesize
\begin{tabular}{lrrl}
\toprule
Partition & Voyage legs & Observations & Use \\
\midrule
Training & 548 & 82{,}472 & Fitting during model selection \\
Validation & 137 & 18{,}753 & Hyperparameter and duration selection \\
Training + validation & 685 & 101{,}225 & Final fit with the selected configuration \\
Test & 176 & 29{,}144 & Evaluation only, latest legs of each vessel \\
\bottomrule
\end{tabular}
\end{table}

\subsection{Evaluation Metrics}
\label{sec:metrics}

Since the mean FOC differs nearly threefold between vessels, absolute errors would be dominated by the larger vessels. We therefore used the overall mean absolute percentage error (MAPE) over all test observations as the primary metric,
\begin{equation}
    \text{MAPE}_{\text{overall}} = \frac{1}{N}\sum_{i=1}^{N} \frac{|y_i - \hat{y}_i|}{y_i} \times 100\%,
\end{equation}
where $N$ is the number of test observations of all vessels. MAPE expresses errors relative to consumption and thus compares vessels with different consumption levels on a common scale. Since the overall MAPE weights each vessel by its number of test observations, we also report the mean per-ship MAPE, that is, the unweighted mean over the 21 vessels. For unseen vessels, we report the median per-ship MAPE instead, because a few poorly predicted vessels would dominate the mean. The coefficient of determination ($R^2$), root mean square error (RMSE), and mean absolute error (MAE) are reported in addition.

\subsection{Implementation of the Compared Methods}
\label{sec:models}

To compare the three modeling approaches of Section~\ref{sec:global_framework} under the same conditions, we implemented each approach with models representative of previous FOC studies. Table~\ref{tab:exp_design} lists the resulting 11 methods, and Fig.~\ref{fig:paradigm_comparison} shows their training procedures. All methods use the same 46 numerical inputs, the same data partition, and the same test legs. The global models other than F2V-Former also receive the ship identity, whereas F2V-Former and the federated and per-ship neural models have no identity input. The architectures and training settings of all methods are given in Appendix~\ref{app:implementation}.

\begin{table}[htbp]
\centering
\caption{The 11 evaluated methods. ``Training data'' is the data seen in the main training phase; ``Per-ship stage'' indicates whether a separate per-ship fine-tuning or training step follows. $N_{\text{model}}$: number of models used at prediction time.}
\label{tab:exp_design}
\footnotesize
\begin{tabular}{llllcr}
\toprule
Method & Approach & Base model & Training data & Per-ship stage & $N_{\text{model}}$ \\
\midrule
\textbf{Global LGBM (ours)} & Global & LightGBM & All fleet (pooled) & None & 1 \\
Global CatBoost & Global & CatBoost & All fleet (pooled) & None & 1 \\
TabM-FiLM & Global & TabM & All fleet (pooled) & None & 1 \\
Global ANN & Global & ANN $[256,128,64]$ & All fleet (pooled) & None & 1 \\
F2V-Former & Global & Transformer (F2V) & All fleet (pooled) & None & 1 \\
\midrule
ANN Transfer & Transfer & ANN $[256,128,64]$ & All fleet $\to$ per-ship FT & Fine-tuning & 21 \\
F2V-Former + LoRA & Transfer & Transformer (F2V) & All fleet $\to$ per-ship LoRA & LoRA & 21 \\
FedAvg + Fine-Tune & Federated & ANN $[256,128,64]$ & Federated $\to$ per-ship FT & Fine-tuning & 21 \\
FedAvg (no fine-tune) & Federated & ANN $[256,128,64]$ & Federated (50 rounds) & None & 1 \\
\midrule
Per-Ship LGBM & Per-ship & LightGBM & Per-ship only & Independent & 21 \\
Per-Ship ANN & Per-ship & ANN $[128,64,32]$ & Per-ship only & Independent & 21 \\
\bottomrule
\end{tabular}
\end{table}

We implemented the global model with five model families to examine whether the results depend on the learning algorithm. Global LightGBM, the primary model, uses gradient-boosted trees \citep{ke2017lightgbm} with categorical splits on the ship identity, and Global CatBoost encodes the ship identity with its own ordered target statistics \citep{prokhorenkova2018catboost}. The Global ANN and TabM-FiLM represent the ship identity by learned parameters, namely a 16-dimensional embedding \citep{j2_entity_embeddings} in the Global ANN, a feed-forward network with hidden layers of 256, 128, and 64 units, and FiLM layers \citep{perez2018film} that scale and shift intermediate representations in TabM-FiLM \citep{k6_tabm}. The fifth implementation is F2V-Former \citep{zhou2026f2vformer}, a Transformer that describes each vessel only by its ship specifications through conditional layer normalization and cross-attention. The hyperparameters of Global LightGBM, Global CatBoost, and TabM-FiLM were selected with 30 Optuna trials each \citep{akiba2019optuna}, whereas the Global ANN uses a fixed configuration and also serves as the pretrained model of ANN Transfer. F2V-Former was selected from 13 configurations on the validation legs and serves as the pretrained model of F2V-Former + LoRA.

Global LightGBM was trained directly on MAPE.\label{sec:mape_objective} Possible training losses include the absolute error, the absolute error weighted by the inverse of the observed FOC, and the absolute percentage error itself. Since the mean FOC differs nearly threefold between vessels, the absolute error emphasizes the larger vessels, whereas the absolute percentage error weighs each observation relative to its consumption and coincides with the evaluation metric. We therefore implemented it as a custom LightGBM objective with the gradient $g_i$ and curvature $h_i$
\begin{equation}
    g_i = \frac{\operatorname{sign}(\hat{y}_i-y_i)}{\tilde y_i},\qquad
    h_i = \operatorname{clip}\!\left(\frac{2}{\tilde y_i\max(|y_i-\hat{y}_i|,\epsilon)},10^{-6},10^4\right),
    \quad \tilde y_i=\max(y_i,0.1),
\end{equation}
where $\epsilon=10^{-6}$ and the lower bound of 0.1\,t/h on $y_i$ does not affect any retained observation. The true second derivative of the absolute percentage error is zero almost everywhere, and $h_i$ is thus a positive surrogate used for numerical optimization and clipped for stability. Validation and early stopping use the exact percentage error. Global CatBoost was trained with its native MAPE loss, and the three neural networks with an L1 loss.

The per-ship baselines train a separate model on the records of each vessel, as shown in Fig.~\ref{fig:paradigm_comparison}d. Per-Ship LightGBM uses the same algorithm and direct MAPE objective as Global LightGBM, with ten Optuna trials per vessel, so that its comparison with Global LightGBM isolates the effect of pooling. Gradient-boosted trees are also a standard choice for per-ship FOC models \citep{b2_comparison, l4_comprehensive_eval}. Per-Ship ANN is a feed-forward network with hidden layers of 128, 64, and 32 units, the per-ship neural baseline used in transfer-learning studies \citep{a1_ann_transfer}.

The adaptation-based baselines learn a model from all vessels and, except for plain FedAvg, fine-tune it on each vessel, as shown in Fig.~\ref{fig:paradigm_comparison}b and c. ANN Transfer follows the pretraining and fine-tuning strategy \citep{pan2010survey, yosinski2014transferable} used by \citet{a1_ann_transfer}, with the Global ANN as the pretrained model whose last hidden layer, output layer, and ship embedding are fine-tuned on the records of each vessel. Since the pretrained model has already seen every vessel, this baseline tests whether adaptation improves a global model for in-fleet vessels. F2V-Former + LoRA applies the same test to a Transformer and adapts the pooled F2V-Former to each vessel by low-rank adaptation of all its linear layers, as in \citet{zhou2026f2vformer}. FedAvg trains the same network without the identity embedding by federated averaging \citep{mcmahan2017fedavg}, as in previous maritime studies \citep{federated_green2023, a5_federated, giannopoulos2024maritime}. Since averaging parameters across vessels with different data distributions is known to degrade federated models \citep{li2020fedprox, li2021fedbn, kairouz2021advances}, FedAvg + Fine-Tune adds a vessel-specific fine-tuning step, a simple form of the personalization used in federated transfer and personalized federated learning \citep{ftl_sm2025, han2024pfl, han2025marl_twin, pan2026ppfl}. The federated training was simulated with centralized preprocessing and does not reproduce a privacy-preserving deployment.

We re-implemented F2V-Former because it is the most recent transfer model evaluated without post-event inputs. The original model reads 14-day sequences of noon reports with 15 static attributes. Our implementation instead reads each 3-hour record together with the inputs of its seven preceding records in the same voyage leg, uses the same 46 inputs as the other methods, and takes the seven ship specifications as its static attributes (Appendix~\ref{app:implementation}). We did not re-implement the transfer model of \citet{a2_transfer_scarcity}, because its inputs include shaft RPM. Whether adaptation helps a vessel absent from training is examined with the few-shot tests in Section~\ref{sec:analyses} for both LightGBM and F2V-Former.

\subsection{In-Fleet Comparison}
\label{sec:infleet}

The in-fleet comparison addresses the main question of this study, namely whether one global model predicts the later voyages of the fleet more accurately than per-ship and adaptation-based models. It evaluates the 11 methods on the test legs of the 21 vessels, each of which is included in training. To confirm that the differences between the methods are not due to chance, we computed 95\% confidence intervals of the differences in overall MAPE between Global LightGBM and each other method with a voyage-leg bootstrap stratified by vessel, which resamples whole test legs within each vessel 10{,}000 times. We also report the refit without the records of other vessels dated within the test period of each target vessel, as described in Section~\ref{sec:eval_protocol}.

In this historical evaluation, the recorded speed, draft, and course and the short-range forecasts matched to the realized track stand in for the voyage plan and its forecasts. Since planned values and longer-range forecasts can deviate from these inputs, we applied the trained global and Per-Ship LightGBM models to perturbed copies of the test inputs, following Monte Carlo analyses of weather uncertainty \citep{mahmoodi2025weather}. Wind speed, wave height, wave period, and current speed were multiplied by $1+\varepsilon$ with $\varepsilon \sim \mathcal{N}(0,s^2)$ and $s \in \{10, 20, 30\}\%$, their directions were shifted by $\mathcal{N}(0,d^2)$ with $d \in \{10^\circ, 20^\circ, 30^\circ\}$, and the speed over ground received additive noise $\mathcal{N}(0,\sigma^2)$ with $\sigma \in \{0.25, 0.5, 1.0\}$\,kn. The noise was drawn either per record or once per voyage leg, with five draws per scenario, and the derived features were recomputed from the perturbed inputs. Besides the record-level MAPE, we report the leg-total error, that is, the absolute percentage error of the total FOC of each test leg, averaged over the 176 test legs.

The benefit of pooling may also differ between vessels. We therefore computed for each vessel $k$ the pooling gain $\Delta_k = \text{MAPE}_k^{\text{per-ship}} - \text{MAPE}_k^{\text{global}}$, that is, the difference in percentage points between the test MAPE of Per-Ship LightGBM and that of Global LightGBM, so that positive values favor the global model. Its Spearman rank correlation with the number of training observations of each vessel indicates whether vessels with fewer records gain more from pooling.

\subsection{Zero-Shot and Few-Shot Prediction}
\label{sec:analyses}

The global model can also predict the FOC of a vessel that has just joined the fleet. This capability is important in practice because the voyages of a newly delivered vessel are planned from its first departure, when no per-ship model can yet be trained and vessel-specific adaptation has no records to use. We therefore evaluated the global model in two settings, namely zero-shot prediction, in which none of the FOC records of the vessel are available, and few-shot prediction, in which only its first records are available.

Zero-shot prediction was evaluated with leave-one-ship-out (LOSO) tests, which treat each vessel in turn as an unseen vessel. For each vessel $k$, all of its records were removed from training and validation, and Global LightGBM was fitted to the other 20 vessels with a fixed configuration (500 rounds, Appendix~\ref{app:implementation}), because tuning on vessel $k$ would require its records. The model then predicted the original test legs of vessel $k$ from its ship specifications and operating inputs, with its ship identity, which the model had not seen in training, treated as missing. F2V-Former was tested in the same way, since it describes each vessel only by its ship specifications, with the configuration and training length selected in the in-fleet comparison and three random seeds. The other neural implementations have no representation for an unknown identity and were therefore not used. As in the in-fleet comparison, we also repeated the tests without the records of other vessels dated within the test period of vessel $k$.

The zero-shot accuracy may also depend on how similar an unseen vessel is to the training fleet. We therefore computed the specification distance $d_{\text{NN}}$ of each vessel, that is, the Euclidean distance to its nearest other vessel in the space of the seven ship specifications, each standardized by the mean and standard deviation over the 21 vessels. Since $d_{\text{NN}}$ uses no FOC value, it can be computed before a vessel enters service. Its association with the zero-shot MAPE was measured with Spearman's rank correlation.

Once an unseen vessel has records, vessel-specific adaptation becomes possible. The few-shot tests examine whether it improves on pooling, using the first $n$ records of vessel $k$ in time order, where $n = 100$, 200, 400, and 800 corresponds to about one to eight median voyage legs. These records were used in three ways, namely added to the pooled data before the global model was refitted with the same configuration (pooled refit), used to continue boosting the zero-shot model (adaptation), or used alone to fit a per-ship model. The adapted and per-ship models selected their number of boosting rounds on the last 20\% of the $n$ records, and all variants were evaluated on the original test legs of vessel $k$. F2V-Former was tested with the pooled refit and with low-rank adaptation of its zero-shot model, whose number of epochs was selected in the same way.

\subsection{Contributing Factors}
\label{sec:factors_method}

The comparisons of Sections~\ref{sec:infleet} and~\ref{sec:analyses} measure how accurate the global model is but not what makes it accurate. We therefore carried out ablation experiments to identify the inputs and the training choice on which its accuracy depends, and examined the trained model with SHAP values.

For in-fleet vessels, we varied one factor at a time. A feature ladder added the ship identity, the ship specifications, and the cross-ship ratios in turn to the 34 base features, that is, the operating, environmental, and within-observation physics features, while the direct MAPE objective was kept fixed. An objective ladder then compared the absolute error, the absolute error weighted by $1/y_i$, and the direct MAPE objective with all 47 features.

Since the ship identity carries no information about an unseen vessel, the inputs that support zero-shot prediction may differ from those that matter for in-fleet vessels. We therefore repeated the LOSO tests of Section~\ref{sec:analyses} with five feature configurations that include the ship identity, ranging from the raw operating and environmental inputs (F5) to all 47 features (F1), and with two of them also without the identity (F2$^{-}$ and F4$^{-}$), giving seven configurations in total. All ablation runs used fixed hyperparameters with row subsampling (fraction 0.8) and were repeated with five random seeds, except the single-seed reference F5, so that the differences between configurations can be compared with the variation between seeds.

We also examined whether the global model relies on physically plausible relationships. The SHAP values \citep{lundberg2017shap} of Global LightGBM on the 29{,}144 test observations rank the features by their mean absolute contribution, and partial dependence plots show how the predicted FOC changes with speed, draft, and wave height.

\subsection{Quarterly Retraining}
\label{sec:retraining_method}

In operation, the global model would be retrained as new records accumulate, because the FOC of a vessel changes over time with hull fouling, maintenance, and operating patterns. We therefore examined whether the model should be retrained and how much past data it should use. In a retrospective analysis over nine quarters (2023\,Q3 to 2025\,Q3), a static Global LightGBM trained on the data up to 2023\,Q2 was compared with models retrained every quarter, either on all past data (cumulative) or on the latest four or two quarters (sliding windows). Each model was evaluated on the quarter following its training data, and the hyperparameters were taken from the main experiment. Since the observations were assigned to calendar quarters, a voyage leg that crosses a quarter boundary can be split between a training and an evaluation quarter.

\section{Results and Discussion}
\label{sec:results}

\subsection{In-Fleet Comparison}
\label{sec:results_comparison}
\label{sec:perturbation}
\label{sec:transfer}

\begin{figure}[htbp]
    \centering
    \includegraphics[width=0.9\textwidth]{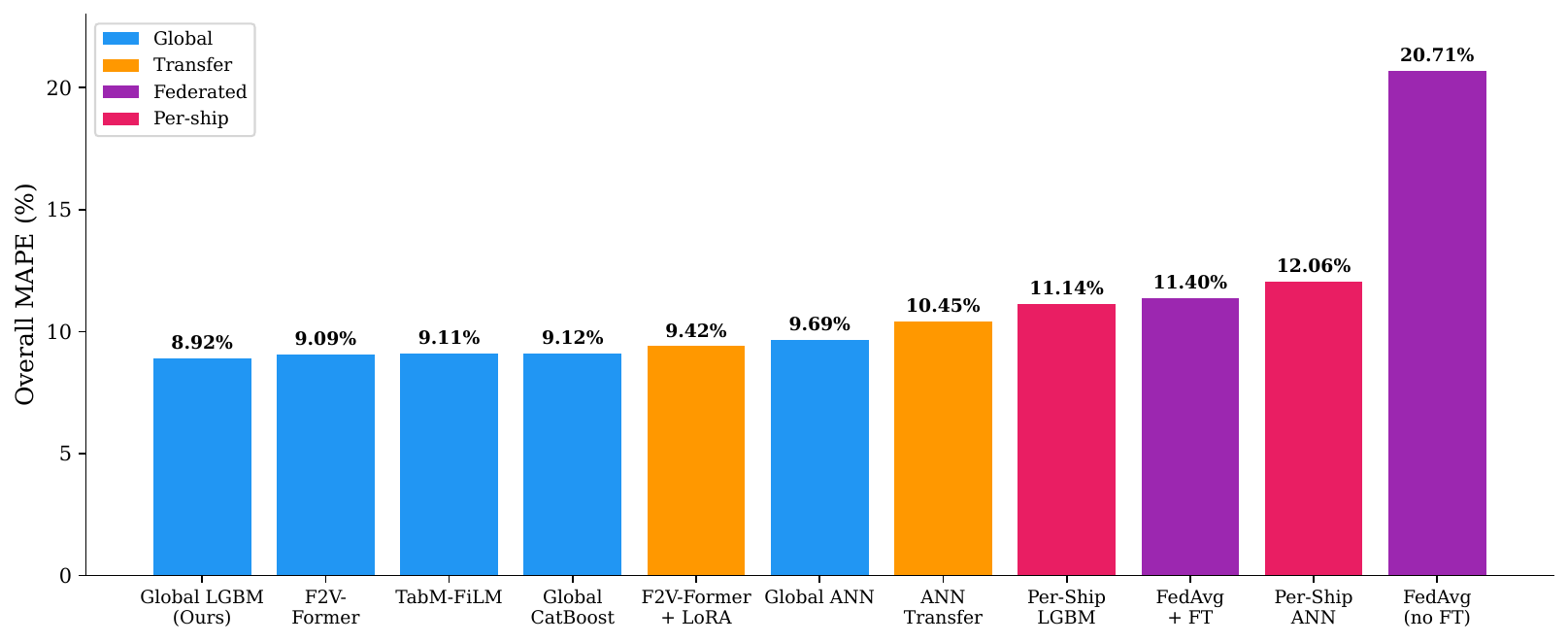}
    \caption{Overall MAPE comparison across all 11 methods, sorted in ascending order. Global LightGBM outperforms all other methods, and every global model outperforms the per-ship and federated models.}
    \label{fig:method_comparison}
\end{figure}

The in-fleet comparison examines whether one global model predicts the later voyages of the fleet more accurately than per-ship and adaptation-based models (Section~\ref{sec:infleet}). Global pooling gives the most accurate predictions for the vessels of the fleet. Table~\ref{tab:method_comparison} and Fig.~\ref{fig:method_comparison} show that Global LightGBM outperforms every per-ship and adaptation-based model and that every global model outperforms the per-ship and federated models. This advantage remains under resampling, without the records of other vessels from the test period, and with perturbed inputs.

Global LightGBM reaches an overall MAPE of 8.92\%, compared with 9.42\% for the best transfer model, F2V-Former + LoRA, 11.14\% for the best per-ship model, and 11.40\% for the best federated model. Even the least accurate global model, the Global ANN at 9.69\%, is more accurate than every per-ship and federated model. The five global models lie within 0.77\,pp of one another, so the advantage does not depend on the learning algorithm. Global LightGBM has the lowest overall MAPE, whereas TabM-FiLM has the lowest mean per-ship MAPE and RMSE and the highest $R^2$, and F2V-Former has the lowest MAE.

\begin{table}[htbp]
\centering
\caption{Method comparison on the temporal test set (29{,}144 observations across 21 vessels). Methods are grouped by approach: global models (no per-ship stage), the adaptation-based transfer and federated models (pretraining $\pm$ per-ship fine-tuning), and per-ship models. Best in bold.}
\label{tab:method_comparison}
\footnotesize
\begin{tabular}{llccccc}
\toprule
Method & Approach & MAPE$_{\text{overall}}$ & MAPE$_{\text{mean}}$ & $R^2$ & RMSE & MAE \\
\midrule
\textbf{Global LGBM (ours)} & Global & \textbf{8.92\%} & 7.97\% & 0.774 & 0.228 & 0.168 \\
F2V-Former & Global & 9.09\% & 8.01\% & 0.790 & 0.219 & \textbf{0.162} \\
TabM-FiLM & Global & 9.11\% & \textbf{7.94\%} & \textbf{0.794} & \textbf{0.218} & 0.163 \\
Global CatBoost & Global & 9.12\% & 8.24\% & 0.790 & 0.220 & 0.165 \\
Global ANN & Global & 9.69\% & 8.71\% & 0.757 & 0.236 & 0.175 \\
\midrule
F2V-Former + LoRA & Transfer & 9.42\% & 8.24\% & 0.760 & 0.235 & 0.169 \\
ANN Transfer & Transfer & 10.45\% & 9.79\% & 0.710 & 0.258 & 0.188 \\
FedAvg + Fine-Tune & Federated & 11.40\% & 11.29\% & 0.691 & 0.266 & 0.201 \\
\midrule
Per-Ship LGBM & Per-ship & 11.14\% & 12.50\% & 0.596 & 0.305 & 0.213 \\
Per-Ship ANN & Per-ship & 12.06\% & 18.31\% & 0.087 & 0.458 & 0.235 \\
FedAvg (no fine-tune) & Federated & 20.71\% & 20.37\% & 0.125 & 0.448 & 0.361 \\
\bottomrule
\end{tabular}
\end{table}

\begin{figure}[htbp]
    \centering
    \includegraphics[width=\textwidth]{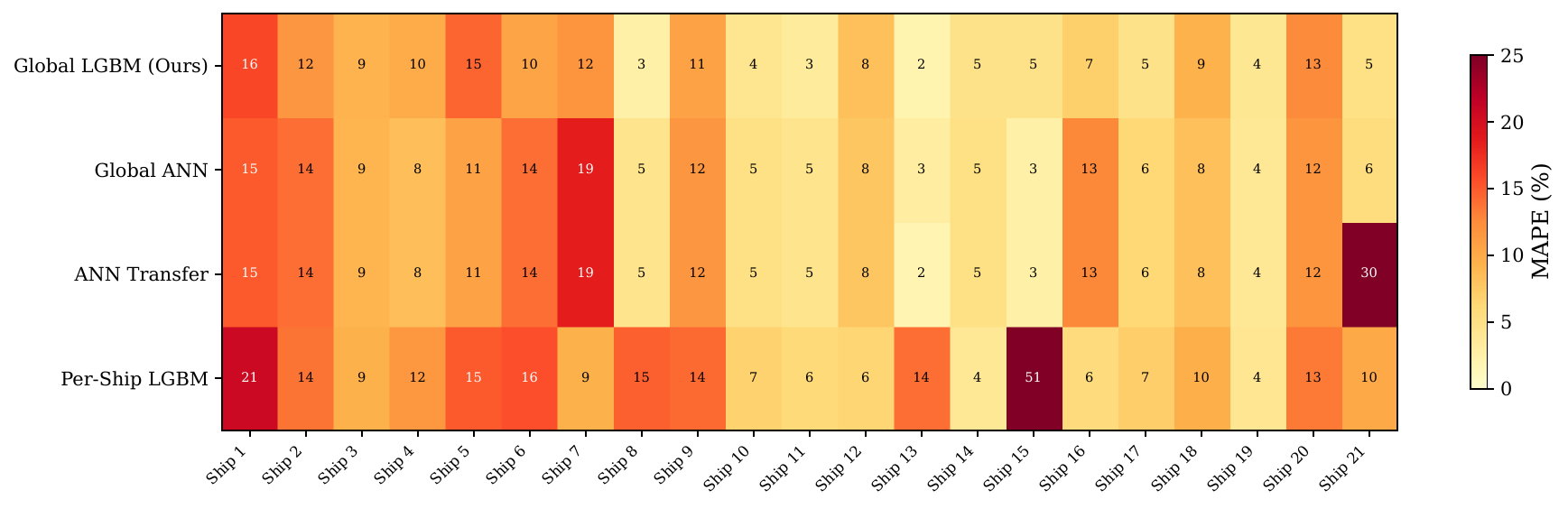}
    \caption{Per-ship MAPE heatmap for four key methods across 21 vessels. Global LightGBM achieves lower MAPE than Per-Ship LightGBM on 17 of 21 vessels, with the largest gains among data-scarce vessels.}
    \label{fig:pership_comparison}
\end{figure}

The advantage is larger when every vessel is weighted equally. The mean per-ship MAPE is 7.97\% for Global LightGBM, compared with 12.50\% for Per-Ship LightGBM and 18.31\% for Per-Ship ANN. Fig.~\ref{fig:pership_comparison} shows that this difference comes mainly from vessels with few records, on which the per-ship models fail. Per-Ship ANN, for example, reaches 106.8\% on Ship~8 and 72.7\% on Ship~13, whereas Global LightGBM reaches 2.6\% and 2.0\% on the same vessels.

Vessel-specific adaptation helps only a model that cannot distinguish the vessels, and even then it does not reach the accuracy of global pooling. Fine-tuning the Global ANN, which already receives the ship identity, does not improve it. Validation selected zero fine-tuning epochs for 19 of the 21 vessels, and fine-tuning degraded Ship~21 from 5.64\% to 29.68\%, so that ANN Transfer reaches 10.45\% against 9.69\% for the Global ANN. Low-rank adaptation of F2V-Former, which describes the vessels by their specifications, does not improve it either. It raises the overall MAPE from 9.09\% to 9.42\%, and from 9.12\% to 9.34\% on average over three seeds, improving nine vessels and degrading nine. The adaptation lowered the validation MAPE from 7.23\% to 6.33\%, but the validation legs lie within the training period, whereas the test legs come later. Fine-tuning the averaged FedAvg model, which has no identity input, reduces its error from 20.71\% to 11.40\%, but FedAvg + Fine-Tune remains less accurate than every global model.

Most of these differences are reliable and do not depend on particular test records. The voyage-leg bootstrap of Section~\ref{sec:infleet} draws the test legs of each vessel at random with replacement, recomputes the differences in overall MAPE, and repeats this 10{,}000 times, so that its 95\% interval shows how much a difference could change with another set of test voyages. The intervals of the differences between Global LightGBM and all per-ship, federated, and ANN Transfer models lie above zero, for example 2.22\,pp (1.59--2.85) for Per-Ship LightGBM. The interval for F2V-Former + LoRA, 0.50\,pp ($-$0.10 to 1.15), narrowly includes zero, as do those for TabM-FiLM, Global CatBoost, and F2V-Former.\footnote{The differences are 1.53\,pp (0.88--2.21) for ANN Transfer, 2.48\,pp (1.66--3.38) for FedAvg + Fine-Tune, 0.18\,pp for TabM-FiLM, and 0.20\,pp for Global CatBoost. The Global ANN is also 0.75\,pp (0.32--1.24) more accurate than ANN Transfer. Low-rank adaptation raises the error of F2V-Former by 0.33\,pp ($-$0.03 to 0.69), and F2V-Former itself is 0.17\,pp less accurate than Global LightGBM.} The advantage also does not come from records of other vessels dated within the test period of a target vessel, since removing these records raises the overall MAPE of Global LightGBM only from 8.92\% to 9.06\%. Nor does it depend on a single vessel. Without Ship~15, on which Per-Ship LightGBM fails with an error of 50.75\%, the overall MAPE is still 8.99\% for the global model and 10.45\% for the per-ship model.

The advantage also persists when the inputs contain plan and forecast errors. Table~\ref{tab:input_noise} shows that, under the largest environmental perturbation (30\% relative noise and $30^\circ$ direction noise), the overall MAPE of Global LightGBM rises from 8.92\% to 9.64\%, and under a 0.5-knot speed error to 9.65\%. Per-Ship LightGBM degrades by similar amounts, so the gap of 2.22\,pp with recorded inputs stays between 1.84 and 2.17\,pp under every perturbation and narrows to 1.21\,pp only for a 1-knot speed error. Errors drawn once per voyage leg, which do not average out within a leg, give gaps of 2.04 and 2.05\,pp. When accuracy is measured on the total FOC of each of the 176 test legs, the error averages 9.32\% for Global LightGBM and 9.13\% for F2V-Former. The five global models (9.13--9.91\%) remain below every per-ship, federated, and ANN Transfer model (10.39--22.46\%). F2V-Former + LoRA lies within the same range at 9.91\%. The median leg-total errors are closer (5.06--7.95\% for all methods except FedAvg), so the advantage comes mainly from legs with large errors. Input errors of the tested size therefore do not change the ranking of the approaches. The perturbations are synthetic, however, and do not reproduce the error structure of operational multi-day forecasts.

\begin{table}[htbp]
\centering
\caption{Robustness of the global advantage to input perturbations (MAPE, \%, mean of five draws; standard deviation across draws $\leq 0.4$\,pp). Environment: wind speed, wave height, wave period, and current speed scaled by $1+\varepsilon$, $\varepsilon \sim \mathcal{N}(0, s^2)$, with $\mathcal{N}(0, d^2)$ added to their directions. Speed: $\mathcal{N}(0, \sigma^2)$ added to speed over ground. Draws are independent per record unless marked per leg. Leg total: mean absolute percentage error of each test leg's total FOC (176 legs).}
\label{tab:input_noise}
\footnotesize
\setlength{\tabcolsep}{4pt}
\begin{tabular}{lccccc}
\toprule
 & \multicolumn{3}{c}{Record MAPE} & \multicolumn{2}{c}{Leg-total error} \\
\cmidrule(lr){2-4}\cmidrule(lr){5-6}
Perturbation & Global LGBM & Per-ship LGBM & Gap (pp) & Global LGBM & Per-ship LGBM \\
\midrule
None (recorded inputs) & 8.92 & 11.14 & 2.22 & 9.32 & 10.97 \\
\addlinespace[2pt]
Environment, $s=10\%$, $d=10^\circ$ & 9.01 & 11.18 & 2.17 & 9.33 & 10.98 \\
Environment, $s=20\%$, $d=20^\circ$ & 9.28 & 11.35 & 2.07 & 9.42 & 11.05 \\
Environment, $s=30\%$, $d=30^\circ$ & 9.64 & 11.58 & 1.94 & 9.55 & 11.17 \\
\addlinespace[2pt]
Speed, $\sigma=0.25$\,kn & 9.10 & 11.25 & 2.15 & 9.17 & 10.98 \\
Speed, $\sigma=0.5$\,kn & 9.65 & 11.60 & 1.94 & 9.14 & 11.01 \\
Speed, $\sigma=1.0$\,kn & 11.58 & 12.79 & 1.21 & 9.24 & 11.36 \\
\addlinespace[2pt]
Environment 20\%/$20^\circ$ + speed 0.5\,kn & 9.99 & 11.83 & 1.84 & 9.46 & 11.34 \\
\addlinespace[2pt]
Per leg: environment 20\%/$20^\circ$ & 9.33 & 11.37 & 2.04 & 9.52 & 11.16 \\
Per leg: speed 0.5\,kn & 9.58 & 11.64 & 2.05 & 9.71 & 11.59 \\
\bottomrule
\end{tabular}
\end{table}

\begin{figure}[htbp]
    \centering
    \includegraphics[width=0.8\textwidth]{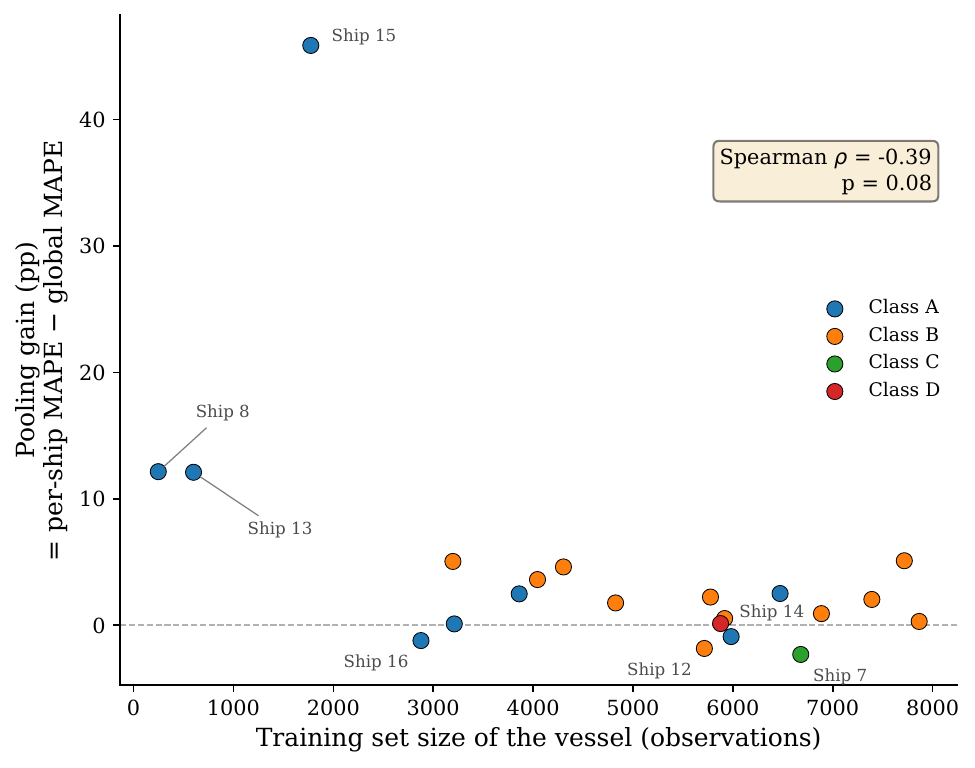}
    \caption{Pooling gain (per-ship MAPE $-$ global MAPE) of each vessel against its training set size, colored by vessel class. Positive values favor the global model. The three smallest training sets receive the three largest gains, and the four labeled vessels below zero are served better by their per-ship models. The fleet-wide rank correlation is weak (Spearman $\rho = -0.39$, $p = 0.08$).}
    \label{fig:transfer_scatter}
\end{figure}

Pooling improves the accuracy for most vessels and most strongly for those with few records. Fig.~\ref{fig:transfer_scatter} shows that the global model is more accurate for 17 of the 21 vessels, with gains of up to 45.9\,pp, whereas the four vessels it predicts less accurately lose at most 2.3\,pp. The three largest gains belong to the vessels with the fewest training records, namely Ship~15 (1{,}774 observations, 4.88\% for the global model versus 50.75\% for the per-ship model, a gain of 45.87\,pp), Ship~8 (247 observations, 2.60\% versus 14.74\%, 12.14\,pp), and Ship~13 (600 observations, 1.96\% versus 14.05\%, 12.10\,pp). Across the fleet, however, the rank correlation between training size and pooling gain is weak (Spearman $\rho = -0.39$, $p = 0.08$), and mid-sized vessels lie on both sides of zero. The vessels with negative gains mostly have many records, namely 5{,}700--6{,}700 training observations for three of the four, so their per-ship models can learn vessel-specific patterns that the global model represents less precisely. Ship~16, with 2{,}877 observations, does not fit this explanation.

For voyage planning, the global model is thus most useful where a per-ship model is weakest. A per-ship model fails when the history of its vessel does not cover the test conditions, as for Ship~15, whereas the global model also learns from the records of other vessels under such conditions. Since the vessels predicted less accurately lose little, an operator does not need to decide vessel by vessel whether enough records exist for a dedicated model.

\subsection{Zero-Shot and Few-Shot Prediction}
\label{sec:loso}

The zero-shot and few-shot tests examine whether the global model can predict the FOC of a vessel that has just joined the fleet, first without any of its records and then with only its first records (Section~\ref{sec:analyses}). Global pooling also predicts most of these vessels with an accuracy comparable to that of per-ship models trained on their records. The zero-shot error increases as the specifications of a vessel differ from those of the other vessels. Once the first records are available, adding them to the pooled data is as accurate as vessel-specific adaptation.

For most vessels, the zero-shot error is even lower than the error of a per-ship model trained on the records of the vessel. Table~\ref{tab:loso} shows that the zero-shot MAPE is lower than that of Per-Ship LightGBM on the same test legs for 14 of the 21 vessels, with a nearly equal mean per-ship MAPE of 12.48\% against 12.50\%. In addition, 17 of the 21 vessels are predicted with errors below 15\%. However, Per-Ship LightGBM remains more accurate in median per-ship MAPE, at 10.14\% against 11.20\%, and in overall MAPE, at 11.14\% against 13.06\%. This accuracy does not depend on records of other vessels dated within the test period of the unseen vessel, since removing these records raises the median per-ship MAPE only from 11.20\% to 12.04\%.

The zero-shot accuracy does not depend on the learning algorithm. F2V-Former, which describes an unseen vessel only by its ship specifications, reaches a median per-ship MAPE of 11.25\% and a mean of 12.33\% over three seeds, against 11.20\% and 12.48\% for Global LightGBM.

\begin{figure}[htbp]
    \centering
    \includegraphics[width=\textwidth]{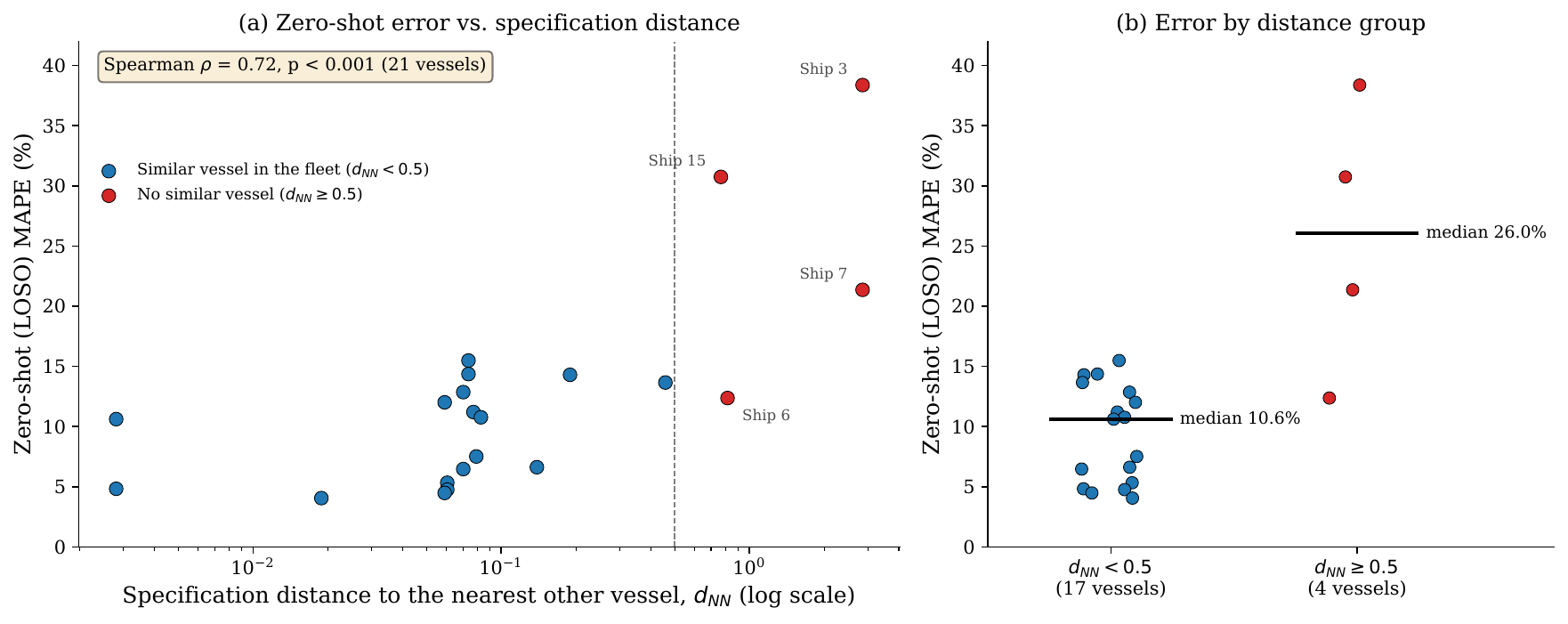}
    \caption{Zero-shot (leave-one-ship-out) error and specification distance. (a)~Zero-shot MAPE of each vessel against its specification distance $d_{\text{NN}}$ to the nearest other vessel (log scale); the four vessels with $d_{\text{NN}} \geq 0.5$ are labeled. (b)~Zero-shot MAPE of the 17 vessels with $d_{\text{NN}} < 0.5$ and of the four vessels with $d_{\text{NN}} \geq 0.5$; bars mark group medians. The 0.5 boundary is descriptive, not an applicability threshold.}
    \label{fig:loso_scatter}
\end{figure}

The zero-shot error depends on whether the fleet contains a vessel with similar specifications. This similarity is measured by the specification distance $d_{\text{NN}}$, that is, the Euclidean distance from a vessel to its nearest other vessel in the space of the seven ship specifications, each standardized over the 21 vessels (Section~\ref{sec:analyses}). A small $d_{\text{NN}}$ thus means that the fleet contains a vessel with nearly the same specifications. Fig.~\ref{fig:loso_scatter} shows that the zero-shot MAPE increases with $d_{\text{NN}}$ (Spearman $\rho = 0.72$, $p < 0.001$).\footnote{For the 19 Class~A and~B vessels, Spearman $\rho=0.63$ and $p=0.004$.} The 17 vessels with $d_{\text{NN}} < 0.5$ reach a median per-ship MAPE of 10.6\%, compared with 26.0\% for the four vessels beyond this distance. Ships~3 and~7, the only oil tanker and the only Class~C bulk carrier of the fleet, have the largest distance of 2.86 and errors of 38.4\% and 21.4\%. Distance alone does not determine the error, however. Ships~15 and~6 lie at similar distances of 0.77 and 0.82 but reach errors of 30.7\% and 12.4\%.

Earlier zero-shot studies did not relate the error of an individual vessel to its similarity to the training fleet. Among LNG carriers of similar size, \citet{parkes2022fleet} found the coverage of operating conditions more important than the vessel parameters. \citet{zhou2026f2vformer} reported similar zero-shot errors across the four capacity classes of a 170-vessel container fleet but compared the errors only by capacity class and route group. The present results show that, in a fleet spanning a 3.7-fold range in deadweight with two single-vessel classes, the specification distance indicates which unseen vessels can be predicted well.

Once the first records of an unseen vessel are available, adding them to the pooled data is as accurate as vessel-specific adaptation and more accurate than a per-ship model trained on them. Table~\ref{tab:new_vessel} shows that, for LightGBM across $n = 100$--800 records, the median per-ship MAPE of the pooled refit and that of adaptation differ by at most 1.4\,pp. Neither method is more accurate on more than 12 of the 21 vessels. The per-ship models fitted on the same records reach a median per-ship MAPE of 15.5--17.5\%, although they are more accurate for a few vessels such as Ship~15. The same holds for F2V-Former. Its pooled refit has a lower mean per-ship MAPE than its low-rank adaptation for every $n$ and a lower median for three of the four values of $n$. The adaptation is also more accurate on only 6 to 9 of the 21 vessels. \citet{zhou2026f2vformer} reported that adapting this model with 500 records of each target vessel reduced the mean MAPE of the target vessels from 13.82\% to 5.64\%. Under the conditions of this study, however, adaptation adds no accuracy to pooling for either model.

The first records improve the prediction mainly for the vessels with large zero-shot errors. After 800 records, the error of Ship~3 falls from 38.4\% to 11.6\% with the pooled refit and to 9.0\% with adaptation. The error of Ship~15 falls from 30.7\% to 10.3\% and 11.2\%. However, the first records raise the error of several vessels that are already predicted well. For Ship~21, the error rises from 6.5\% to 13.0\% with the pooled refit and to 27.2\% with adaptation, possibly because its first records date from 2021, more than two years before its test legs. As a result, the median per-ship MAPE stays near 11\%, whereas the mean falls from 12.48\% to about 11\% at 800 records.

\begin{table}[htbp]
\centering
\caption{Zero-shot prediction in leave-one-ship-out (LOSO) tests for all 21 vessels. $d_{\text{NN}}$: specification distance, that is, the Euclidean distance to the nearest other vessel in standardized specification space (Section~\ref{sec:analyses}). LOSO MAPE: error of Global LightGBM fitted without any records of the vessel. Per-Ship LGBM and Global LGBM: in-fleet errors of Per-Ship LightGBM and Global LightGBM on the same test legs (Table~\ref{tab:method_comparison}). Vessels are sorted by LOSO MAPE. Summary rows give the median and mean per-ship MAPE and the overall MAPE.}
\label{tab:loso}
\footnotesize
\begin{tabular}{lcrrrrr}
\toprule
Ship & Class & $d_{\text{NN}}$ & $n_{\text{test}}$ & LOSO MAPE & Per-Ship LGBM & Global LGBM \\
\midrule
Ship 3   & D & 2.856 & 1{,}530 & 38.37\% & 9.39\% & 9.26\% \\
Ship 15  & A & 0.767 & 501 & 30.74\% & 50.75\% & 4.88\% \\
Ship 7   & C & 2.856 & 1{,}970 & 21.36\% & 9.40\% & 11.72\% \\
Ship 1   & B & 0.074 & 1{,}304 & 15.48\% & 20.67\% & 16.06\% \\
Ship 5   & B & 0.074 & 1{,}631 & 14.36\% & 15.04\% & 14.52\% \\
Ship 14  & A & 0.189 & 1{,}878 & 14.30\% & 3.90\% & 4.81\% \\
Ship 2   & B & 0.458 & 1{,}917 & 13.66\% & 13.69\% & 11.65\% \\
Ship 20  & B & 0.070 & 2{,}089 & 12.86\% & 13.45\% & 12.54\% \\
Ship 6   & B & 0.816 & 2{,}134 & 12.37\% & 15.60\% & 10.50\% \\
Ship 18  & B & 0.059 & 2{,}069 & 12.01\% & 9.62\% & 9.33\% \\
Ship 4   & B & 0.077 & 1{,}523 & 11.20\% & 11.60\% & 9.84\% \\
Ship 12  & B & 0.083 & 1{,}842 & 10.77\% & 6.38\% & 8.23\% \\
Ship 19  & A & 0.003 & 830 & 10.61\% & 4.17\% & 4.07\% \\
Ship 9   & B & 0.079 & 1{,}143 & 7.51\% & 14.26\% & 10.65\% \\
Ship 16  & A & 0.139 & 856 & 6.62\% & 5.75\% & 6.97\% \\
Ship 21  & B & 0.070 & 924 & 6.46\% & 10.14\% & 5.10\% \\
Ship 10  & A & 0.061 & 1{,}009 & 5.33\% & 6.80\% & 4.33\% \\
Ship 11  & A & 0.003 & 1{,}814 & 4.83\% & 5.97\% & 3.47\% \\
Ship 8   & A & 0.061 & 243 & 4.76\% & 14.74\% & 2.60\% \\
Ship 17  & B & 0.059 & 1{,}713 & 4.48\% & 7.06\% & 4.84\% \\
Ship 13  & A & 0.019 & 224 & 4.05\% & 14.05\% & 1.96\% \\
\midrule
\textbf{Median} & & \textbf{0.074} & & \textbf{11.20\%} & 10.14\% & 8.23\% \\
\textbf{Mean}   & & & & \textbf{12.48\%} & 12.50\% & 7.97\% \\
\textbf{Overall} & & & & \textbf{13.06\%} & 11.14\% & 8.92\% \\
\bottomrule
\end{tabular}
\end{table}

\begin{table}[htbp]
\centering
\caption{Few-shot prediction for unseen vessels with their first $n$ records: median (mean) per-ship MAPE (\%) over the 21 vessels, evaluated on the original test legs of each vessel. Pooled refit: the $n$ records are added to the pooled data and the global model is refitted with the fixed configuration of the LOSO tests. Adaptation: the zero-shot model is boosted further on the $n$ records. Per-ship: a model fitted on the $n$ records alone. F2V-Former (seed 42): the $n$ records are added to the pooled data before refitting, or used for low-rank adaptation (LoRA) of its zero-shot model. Single seed. \textsuperscript{$\dagger$}Ship~8 (247 records) and, for $n = 800$, Ship~13 (600 records) use all their records.}
\label{tab:new_vessel}
\footnotesize
\setlength{\tabcolsep}{4pt}
\begin{tabular}{lccccc}
\toprule
 & \multicolumn{3}{c}{LightGBM} & \multicolumn{2}{c}{F2V-Former} \\
\cmidrule(lr){2-4}\cmidrule(lr){5-6}
$n$ (records) & Pooled refit & Adaptation & Per-ship & Pooled refit & LoRA \\
\midrule
0 (zero-shot) & \multicolumn{3}{c}{11.20 (12.48)} & \multicolumn{2}{c}{12.19 (13.01)} \\
100 & 12.88 (12.67) & 11.48 (12.37) & 17.46 (21.99) & 10.65 (10.31) & 12.45 (13.10) \\
200 & 11.59 (12.78) & 11.14 (12.78) & 15.47 (20.20) & 10.33 (9.96) & 10.64 (11.19) \\
400\textsuperscript{$\dagger$} & 10.42 (11.80) & 11.31 (11.85) & 16.70 (18.68) & 10.92 (10.31) & 10.28 (10.99) \\
800\textsuperscript{$\dagger$} & 11.02 (10.95) & 11.22 (11.29) & 17.05 (18.90) & 9.59 (9.51) & 10.67 (10.99) \\
\bottomrule
\end{tabular}
\end{table}

\subsection{Contributing Factors}
\label{sec:ablation}

The analysis of contributing factors examines what makes the global model accurate for in-fleet and unseen vessels and whether it relies on physically plausible relationships (Section~\ref{sec:factors_method}). The accuracy of the global model depends mainly on how each vessel is described. Ship identity and ship specifications are largely interchangeable for in-fleet vessels, but only the specifications describe an unseen vessel. Matching the training objective to MAPE adds a further 0.64\,pp. The relationships that the model has learned between the inputs and FOC are physically plausible.

For in-fleet vessels, ship identity and ship specifications carry largely the same information. Table~\ref{tab:ablation} shows that the overall MAPE falls from 12.00\% to 9.75\% when the ship identity is added to the 34 base features, to 9.16\% when the ship specifications are added, and to 9.06\% when the cross-ship ratios are added. However, Table~\ref{tab:regime_ablation} shows that, once the specifications are included, removing the identity raises the in-fleet MAPE only from 9.16\% to 9.34\%. The large gain from the identity in the first step thus comes mostly from information that the specifications also provide. The full model reaches 9.06\% here rather than the 8.92\% of Section~\ref{sec:results_comparison} because the ablation runs use fixed hyperparameters instead of tuning each configuration.

Matching the training objective to the evaluation metric also improves the accuracy. Table~\ref{tab:ablation} shows that, with all 47 features, the overall MAPE falls from 9.70\% with the absolute error to 9.39\% with the absolute error weighted by $1/y_i$ and to 9.06\% with the direct MAPE objective of Section~\ref{sec:mape_objective}. The direct MAPE objective is thus worth 0.64\,pp over the absolute error.

\begin{figure}[htbp]
    \centering
    \includegraphics[width=\textwidth]{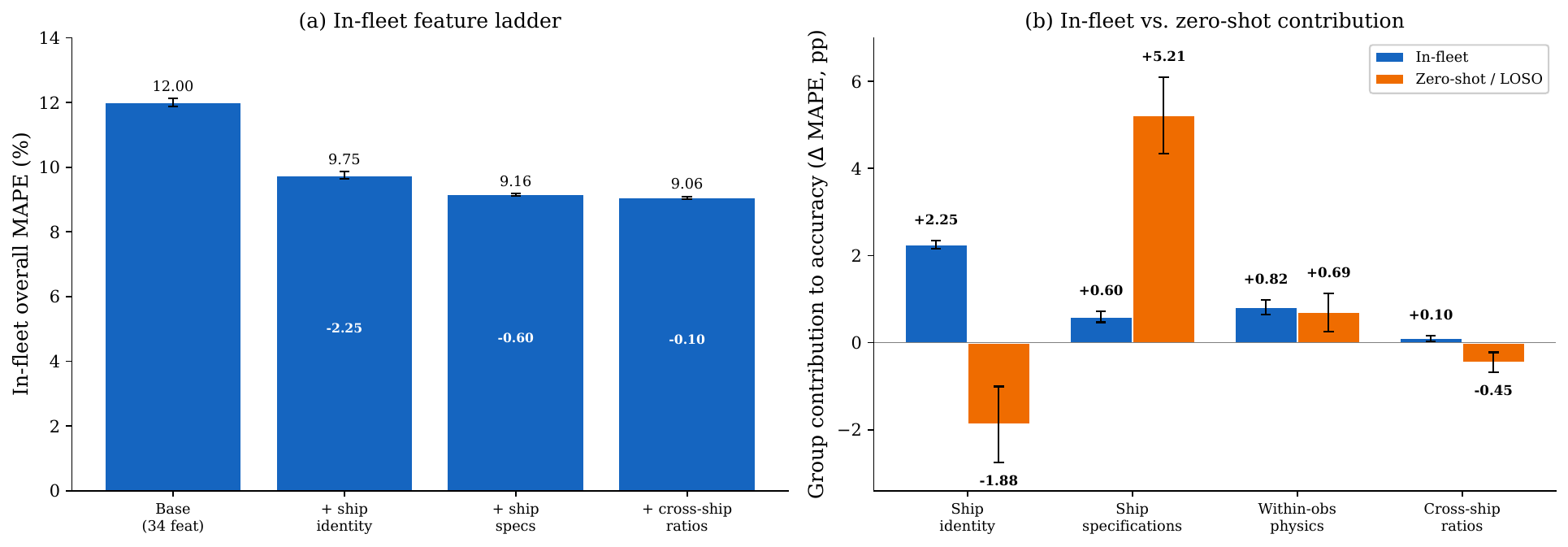}
    \caption{Ablation results. (a)~In-fleet feature ladder: overall MAPE as each feature group is added, training objective fixed at direct MAPE; numbers inside bars give the step change. (b)~Feature-group contribution to accuracy ($\Delta$ MAPE, positive = the group helps), in-fleet versus zero-shot. Both identity bars compare the base features with and without identity; with specifications present (F2 versus F2$^{-}$), identity contributes $+0.19$\,pp in-fleet and $-0.06$\,pp under LOSO. Five seeds; error bars $\pm 1$ standard deviation.}
    \label{fig:ablation}
\end{figure}

For unseen vessels, the ship specifications are the decisive feature group. Table~\ref{tab:regime_ablation} and Fig.~\ref{fig:ablation}b show that adding the specifications to the base features and the ship identity lowers the zero-shot median per-ship MAPE from 16.00\% to 10.78\%, whereas the same step lowers the in-fleet MAPE by only 0.60\,pp. The within-observation physics features and the cross-ship ratios change the zero-shot median per-ship MAPE by at most 0.7\,pp. These zero-shot values differ slightly from those of Section~\ref{sec:loso}, for example 11.24\% against 11.20\% for the full configuration, because the ablation runs use five seeds and row subsampling.

The ship identity does not help unseen vessels and even harms them when the specifications are absent. With the specifications included, the zero-shot median per-ship MAPE is 10.72\% without the identity and 10.78\% with it, a difference smaller than the variation between seeds. Without the specifications, the identity raises the zero-shot median per-ship MAPE from 14.12\% to 16.00\%, although it lowers the in-fleet MAPE by 2.25\,pp in the same comparison. The reason is that the unknown identity of an unseen vessel follows the tree splits learned for other vessels. Without the specifications, the unseen vessel is therefore predicted like the other vessels on the same branch. The identity thus improves the accuracy only for vessels whose records were used in training, whereas the specifications describe any vessel. This result is consistent with the dependence of the zero-shot error on the specification distance in Section~\ref{sec:loso}.

\begin{figure}[htbp]
    \centering
    \includegraphics[width=0.8\textwidth]{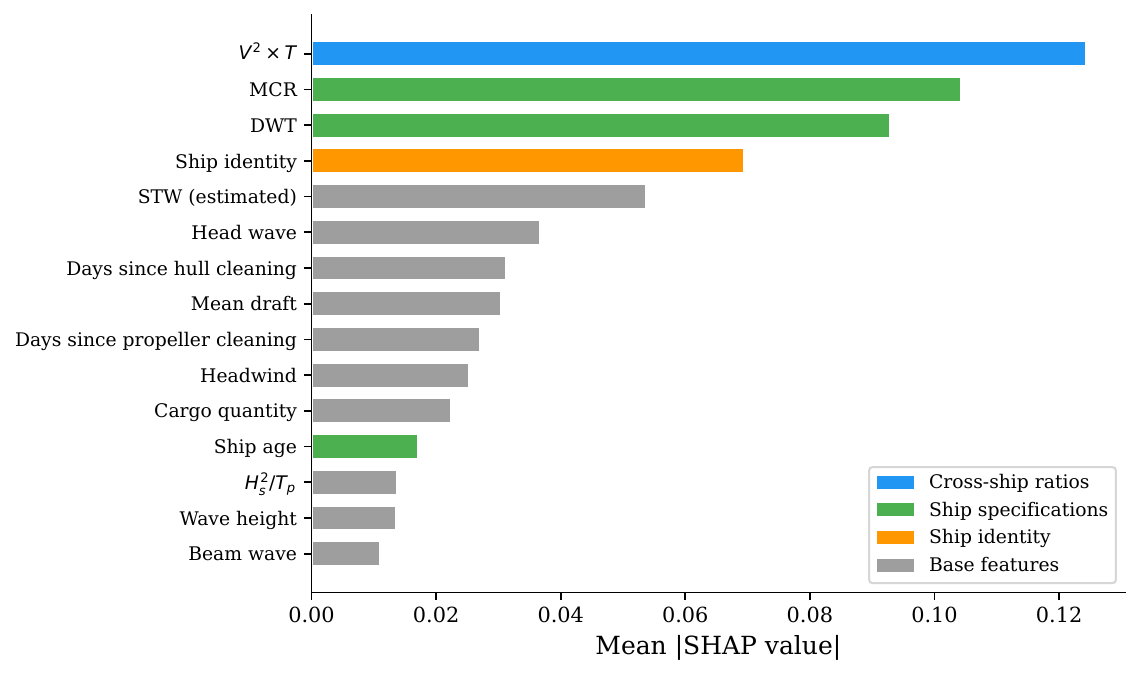}
    \caption{Top 15 features by mean $|\text{SHAP}|$ value. Colors indicate feature groups: blue = cross-ship ratios, green = ship specifications, orange = ship identity, gray = base features.}
    \label{fig:shap}
\end{figure}

The global model relies mainly on the size and engine power of a vessel and on its speed and loading. Fig.~\ref{fig:shap} shows that the five largest mean absolute SHAP values belong to $V^2T$, MCR, DWT, the ship identity, and the estimated speed through water. The days since hull and propeller cleaning rank seventh and ninth, suggesting that the model also tracks the hull and propeller condition over time. However, a high attribution does not mean that a feature is indispensable. Without the cross-ship ratios, including $V^2T$, the in-fleet MAPE rises by only 0.10\,pp because correlated raw and derived inputs share the same information.

\begin{figure}[htbp]
    \centering
    \includegraphics[width=\textwidth]{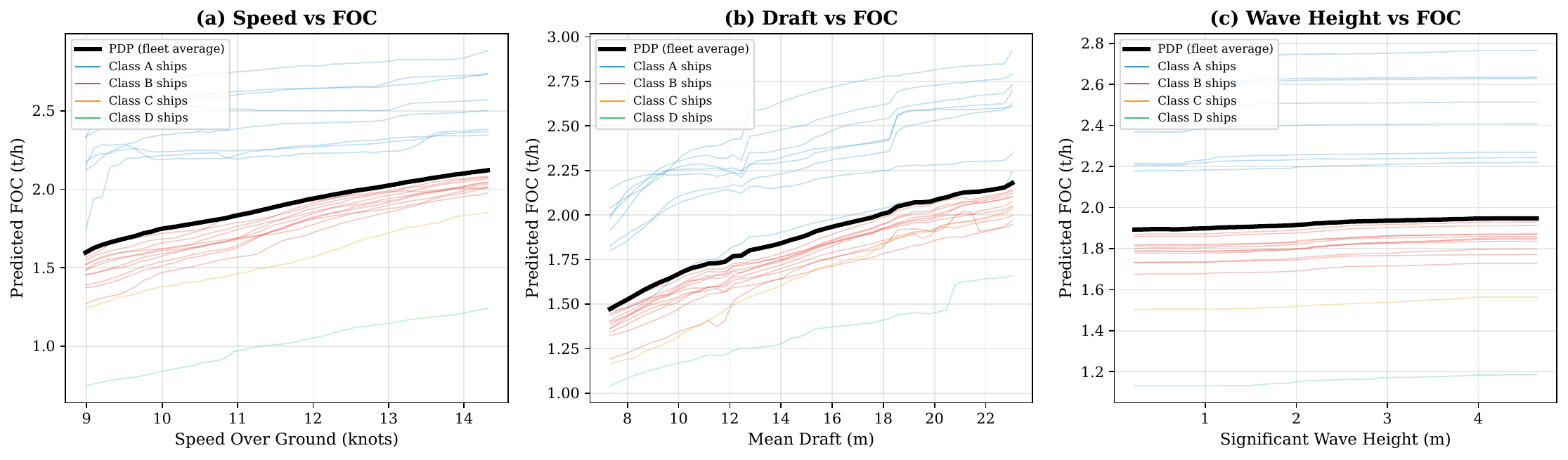}
    \caption{Partial dependence plots: (a) speed $\rightarrow$ FOC (increasing), (b) draft $\rightarrow$ FOC (monotonically increasing), (c) wave height $\rightarrow$ FOC (weakly increasing). The directions are physically plausible.}
    \label{fig:pdp}
\end{figure}

The learned relationships are physically plausible. Fig.~\ref{fig:pdp} shows that the predicted FOC increases with speed, draft, and wave height. The increase with speed is gentler than the cubic law \citep{i4_survey} because the effect of speed is shared among several features, namely the speed, $V^2$, $V^3$, the estimated speed through water, and $V^2T$. The partial dependence on the speed alone therefore understates this effect.

\begin{table}[htbp]
\centering
\caption{In-fleet ablation: overall MAPE, mean $\pm$ standard deviation over five seeds, fixed hyperparameters. $\Delta$: seed-paired change from the preceding row. The 35-, 42-, and 47-feature rows equal F4, F2, and F1 of Table~\ref{tab:regime_ablation}. \textsuperscript{$\dagger$}MAE with sample weights $1/y_i$.}
\label{tab:ablation}
\footnotesize
\begin{tabular}{lccc}
\toprule
Configuration & $n_{\text{feat}}$ & In-fleet MAPE (\%) & $\Delta$ (pp) \\
\midrule
\multicolumn{4}{l}{\emph{Feature ladder (objective fixed: direct MAPE)}} \\
Base features & 34 & 12.00 $\pm$ 0.12 & --- \\
+ ship identity (categorical) & 35 & 9.75 $\pm$ 0.10 & $-$2.25 $\pm$ 0.10 \\
+ ship specifications & 42 & 9.16 $\pm$ 0.04 & $-$0.60 $\pm$ 0.13 \\
+ cross-ship ratios & 47 & \textbf{9.06 $\pm$ 0.04} & $-$0.10 $\pm$ 0.06 \\
\midrule
\multicolumn{4}{l}{\emph{Objective ladder (features fixed: full 47)}} \\
MAE & 47 & 9.70 $\pm$ 0.09 & --- \\
MAPE-weighted MAE\textsuperscript{$\dagger$} & 47 & 9.39 $\pm$ 0.14 & $-$0.31 $\pm$ 0.19 \\
Direct MAPE objective & 47 & \textbf{9.06 $\pm$ 0.04} & $-$0.33 $\pm$ 0.13 \\
\bottomrule
\end{tabular}
\end{table}

\begin{table}[htbp]
\centering
\caption{Zero-shot ablation: feature-group configurations evaluated in-fleet and leave-one-ship-out (LOSO median and mean over the 21 vessels, each predicted as an unseen vessel). Mean $\pm$ standard deviation over five seeds, fixed hyperparameters; F5\textsuperscript{$\dagger$}: single seed. F1--F5 include ship identity; F2$^{-}$ and F4$^{-}$ omit it. \emph{raw} = operational + environmental + maintenance; \emph{within} = within-observation physics (13); \emph{specs} = ship specifications (7); \emph{ratios} = cross-ship ratios (5).}
\label{tab:regime_ablation}
\footnotesize
\setlength{\tabcolsep}{4pt}
\begin{tabular}{llcccc}
\toprule
 & Configuration & $n_{\text{feat}}$ & In-fleet MAPE (\%) & LOSO median (\%) & LOSO mean (\%) \\
\midrule
F5\textsuperscript{$\dagger$} & Raw & 22 & 10.48 & 15.13 & 18.19 \\
F3 & Raw + specs & 29 & 9.97 $\pm$ 0.13 & 11.48 $\pm$ 0.56 & 13.15 $\pm$ 0.20 \\
F4 & Raw + within & 35 & 9.75 $\pm$ 0.10 & 16.00 $\pm$ 0.91 & 17.45 $\pm$ 0.16 \\
F4$^{-}$ & Raw + within, no identity & 34 & 12.00 $\pm$ 0.12 & 14.12 $\pm$ 0.14 & 15.39 $\pm$ 0.11 \\
F2 & Raw + within + specs & 42 & 9.16 $\pm$ 0.04 & 10.78 $\pm$ 0.25 & 12.77 $\pm$ 0.42 \\
F2$^{-}$ & Raw + within + specs, no identity & 41 & 9.34 $\pm$ 0.10 & \textbf{10.72 $\pm$ 0.23} & 12.44 $\pm$ 0.16 \\
F1 & Full (+ ratios) & 47 & \textbf{9.06 $\pm$ 0.04} & 11.24 $\pm$ 0.24 & \textbf{12.21 $\pm$ 0.12} \\
\bottomrule
\end{tabular}
\end{table}

\subsection{Quarterly Retraining}
\label{sec:retraining}

The retraining analysis examines whether the global model should be retrained in operation and how much past data it should use (Section~\ref{sec:retraining_method}). Retraining every quarter lowers the error substantially. At least one year of history is needed, but beyond one year the length of the history matters little.

Retraining every quarter makes the global model substantially more accurate than a static model. Table~\ref{tab:retraining} shows that, averaged over the nine quarters, the overall MAPE is 14.68\% for the static model, against 9.04\% for cumulative retraining, 8.84\% for the four-quarter window, and 9.85\% for the two-quarter window. This difference does not come from the vessels that joined the fleet after the training period of the static model. On the 17 vessels included in its training data, the static model averages 15.07\%, against 9.46\% for cumulative retraining and 9.33\% for the four-quarter window. These averages are not directly comparable with the test MAPE of Section~\ref{sec:results_comparison} because the evaluation periods and partitions differ.

The gain comes from recent records rather than from a larger amount of data. The four-quarter window trains on 22{,}000 to 36{,}000 records, fewer than the 58{,}000 records of the static model, but its average error is 5.84\,pp lower. This result indicates that the FOC of the vessels changes over time in ways that the inputs do not fully capture, for example through hull fouling, maintenance, and changes in operating patterns.

Beyond one year, the length of the history matters little. Cumulative retraining and the four-quarter window differ by only 0.20\,pp on average. The four-quarter window is more accurate in five of the nine quarters and cumulative retraining in the other four. The two-quarter window is less accurate, possibly because six months of records cover fewer operating conditions and in some quarters contain no records of a vessel that the longer histories include.

\begin{table}[htbp]
\centering
\caption{Quarterly retraining results (overall MAPE, \%) of Global LightGBM with the hyperparameters of the main experiment. Static: trained once on the data up to 2023\,Q2. Cumulative: retrained every quarter on all past data. Sliding 4Q/2Q: retrained every quarter on the latest four or two quarters. 2025Q3 contains July 2025 only, the last month of the dataset. Best per quarter in bold.}
\label{tab:retraining}
\footnotesize
\begin{tabular}{lcccc}
\toprule
Quarter & Static & Cumulative & Sliding 4Q & Sliding 2Q \\
\midrule
2023Q3 & 14.87 & 14.87 & \textbf{13.41} & 14.35 \\
2023Q4 & 15.01 & 11.83 & 8.93 & \textbf{8.26} \\
2024Q1 & 15.13 & 9.72 & \textbf{9.42} & 10.41 \\
2024Q2 & 13.86 & \textbf{7.52} & 8.49 & 10.30 \\
2024Q3 & 12.38 & \textbf{7.36} & 8.50 & 9.41 \\
2024Q4 & 14.01 & 7.22 & \textbf{6.76} & 7.39 \\
2025Q1 & 15.19 & \textbf{8.56} & 9.04 & 10.92 \\
2025Q2 & 15.40 & \textbf{8.38} & 9.52 & 11.68 \\
2025Q3 & 16.24 & 5.87 & \textbf{5.47} & 5.94 \\
\midrule
Mean & 14.68 & 9.04 & \textbf{8.84} & 9.85 \\
\bottomrule
\end{tabular}
\end{table}

\subsection{Practical Implications and Limitations}
\label{sec:implications}
\label{sec:limitations}

The results have implications for fleet operators who keep the records of their vessels in one place and for researchers who compare FOC prediction approaches. For operators, one global model can serve both the vessels already in the fleet and newly delivered vessels. For researchers, the evaluation protocol of this study provides a basis for comparing approaches on equal terms.

For the vessels already in the fleet, one global model can replace the per-ship and adapted models. The global model was more accurate overall than every per-ship and adaptation-based model and more accurate than Per-Ship LightGBM for 17 of the 21 vessels. It also removes the per-ship training and the fine-tuning step, so that the operator maintains one model instead of 21. However, the errors of individual vessels should still be monitored, because the global model was less accurate than Per-Ship LightGBM for four vessels, by up to 2.3\,pp.

If the ship specifications are recorded accurately, the global model can also predict a newly delivered vessel from its first voyage. Before the vessel enters service, the operator can compute its specification distance to the fleet. If the fleet contains a vessel with similar specifications, the zero-shot prediction can serve as an initial estimate, since for most vessels its accuracy is comparable to that of a per-ship model trained on the records of the vessel. Otherwise, the prediction should be treated as provisional. Adding the first records of the vessel to the pooled data then corrects large zero-shot errors without fine-tuning. Because these records can also raise the error of a vessel that is already predicted well, the error should be monitored during the first voyages.

In operation, the global model should be retrained regularly with at least one year of history. In the retrospective analysis, quarterly retraining lowered the average error from 14.68\% to between 8.84\% and 9.85\%. A history longer than one year added little, whereas a six-month history was less accurate. The retraining interval and the history length should still be checked for the fleet in which the model is deployed.

For researchers, FOC prediction approaches should be compared under pre-departure conditions. The inputs should be limited to variables known or forecast before departure, such as the planned speed and draft and the weather forecasts. Every compared method should receive the same inputs and be tested on the later, intact voyage legs of each vessel. Otherwise, the reported accuracies are neither comparable between studies nor indicative of the accuracy available for voyage planning.

These implications are drawn from the fleet of a single operator, consisting of 20 bulk carriers and one oil tanker mostly on the Asia--South America trade. Because the vessels share a trade and an operator, the relationship between speed, loading, weather, and FOC is similar across them and thus favors pooling. The zero-shot prediction is also reliable only for vessels similar to those in the fleet. The oil tanker and the Class~C bulk carrier, the only vessels of their classes, had zero-shot errors of 38.4\% and 21.4\%. The advantage of global pooling therefore needs to be confirmed on fleets of several operators and vessel types.

The accuracy reported in this study is likely higher than the accuracy achievable in actual voyage planning. The evaluation used the recorded speed and draft and short-range weather forecasts along the actual track, whereas a voyage plan relies on planned values and forecasts issued several days before departure. When errors were added to these inputs, the overall MAPE of Global LightGBM rose from 8.92\% to at most 11.58\%, but the global model remained more accurate than Per-Ship LightGBM in every case. Archived operational forecasts are needed to measure the accuracy under actual planning conditions.

The accuracy also depends on the maintenance records of each vessel. For the six vessels without records of hull or propeller cleaning, the mean per-ship MAPE of Global LightGBM was 11.92\%, against 6.39\% for the other 15 vessels. Part of this difference may come from other differences between the vessels, since the six vessels include the oil tanker and the Class~C bulk carrier. Nevertheless, the result suggests that complete maintenance records are important for accurate prediction.

The conclusion that vessel-specific adaptation is unnecessary holds for the adaptation methods implemented in this study, which include the low-rank adaptation of the Transformer of \citet{zhou2026f2vformer}. In this study, however, the Transformer was pretrained on at most 21 vessels with seven ship specifications, whereas \citet{zhou2026f2vformer} pretrained it on the source vessels of a 170-vessel fleet with 15 static attributes, including the service route. Adaptation may therefore help more when the pretraining fleet is larger and more diverse.

\section{Conclusion}
\label{sec:conclusion}

This study examined whether FOC prediction for voyage planning requires per-ship modeling or vessel-specific adaptation, or whether one global model trained on the pooled records of a fleet is sufficient. Per-ship, adaptation-based, and global models were compared on 21 physically diverse vessels under pre-departure conditions, in which all methods received the same inputs known or forecast before departure and were tested on the later, intact voyage legs of each vessel. Global pooling performed best overall in every setting examined.

First, global pooling is the most accurate approach for in-fleet vessels. Global LightGBM achieves a lower overall MAPE than every per-ship, transfer, and federated model, with 8.92\% against 11.14\% for Per-Ship LightGBM. Global models of all five architectures, from gradient-boosted trees to a Transformer, outperform the per-ship and federated models. Fine-tuning a global model for each vessel adds no accuracy, even with the low-rank adaptation of a recent Transformer. The advantage holds under resampling of the test voyages, without the records of other vessels from the test period, and with perturbed inputs.

Second, the advantage of global pooling extends to unseen vessels. In the LOSO tests, the global model reaches a median per-ship MAPE of 11.20\% without any records of the vessel, with 17 of the 21 vessels below 15\%. For 14 of these vessels, the zero-shot error is even lower than the error of Per-Ship LightGBM trained on the records of the vessel. The zero-shot error increases with the specification distance of a vessel to the rest of the fleet (Spearman $\rho = 0.72$). Once the first records of an unseen vessel are available, adding them to the pooled data is as accurate as vessel-specific adaptation, for both gradient-boosted trees and the Transformer, and more accurate than per-ship models trained on these records.

Third, the analysis of contributing factors identifies what enables these results. For in-fleet vessels, ship identity and ship specifications carry largely the same information. For unseen vessels, only the specifications support prediction and lower the zero-shot median per-ship MAPE from 16.00\% to 10.78\%. Without the specifications, the identity even harms the prediction of unseen vessels. Matching the training objective to MAPE adds a further 0.64\,pp.

These results show that per-ship models and vessel-specific adaptation are not necessary for accurate FOC prediction once the approaches are compared under fair pre-departure conditions. One global model, retrained regularly as records accumulate, can serve both the vessels of a fleet and newly delivered vessels with specifications similar to those of the fleet. An operator therefore needs to develop and maintain a single model instead of one per vessel. The evaluation protocol of this study excludes information unavailable before departure and thus allows future studies to compare FOC prediction approaches on equal terms.

Future work should test global pooling on fleets of several operators and vessel types and with archived forecasts issued several days before departure. Complete maintenance records would show how much of the remaining error comes from the condition of the hull and propeller. In addition, larger and more diverse pretraining fleets would show whether vessel-specific adaptation becomes useful when the source vessels differ more from the target vessel.

\section*{Acknowledgments}

This work was supported by the National IT Industry Promotion Agency (NIPA) grant funded by the Korea government (S2201-24-1003, Development of analysis engine and API service capable of AI-based ship greenhouse gas management and reduction).

\section*{Data availability}

The operational dataset and the associated code cannot be shared publicly under the confidentiality agreement with the fleet operator. To support methodological reproducibility, Appendix~\ref{app:implementation} documents every method's configuration in full (architectures, hyperparameter search spaces, training protocols, and the federated averaging algorithm), so that each of the 11 methods can be re-implemented on comparable operational data.

\section*{Declaration of Generative AI and AI-assisted technologies in the writing process}

During the preparation of this work, the authors used Claude and OpenAI Codex for language editing and checking manuscript consistency. After using these tools, the authors reviewed and edited the content as needed and took full responsibility for the content of the published article.

\bibliographystyle{elsarticle-harv}
\bibliography{references}

\begin{thebibliography}{53}
\expandafter\ifx\csname natexlab\endcsname\relax\def\natexlab#1{#1}\fi
\providecommand{\url}[1]{\texttt{#1}}
\providecommand{\href}[2]{#2}
\providecommand{\path}[1]{#1}
\providecommand{\DOIprefix}{doi:}
\providecommand{\ArXivprefix}{arXiv:}
\providecommand{\URLprefix}{URL: }
\providecommand{\Pubmedprefix}{pmid:}
\providecommand{\doi}[1]{\href{http://dx.doi.org/#1}{\path{#1}}}
\providecommand{\Pubmed}[1]{\href{pmid:#1}{\path{#1}}}
\providecommand{\bibinfo}[2]{#2}
\ifx\xfnm\relax \def\xfnm[#1]{\unskip,\space#1}\fi
\bibitem[{Akiba et~al.(2019)Akiba, Sano, Yanase, Ohta and
  Koyama}]{akiba2019optuna}
\bibinfo{author}{Akiba, T.}, \bibinfo{author}{Sano, S.},
  \bibinfo{author}{Yanase, T.}, \bibinfo{author}{Ohta, T.},
  \bibinfo{author}{Koyama, M.}, \bibinfo{year}{2019}.
\newblock \bibinfo{title}{{Optuna}: A next-generation hyperparameter
  optimization framework}, in: \bibinfo{booktitle}{Proceedings of the 25th ACM
  SIGKDD International Conference on Knowledge Discovery and Data Mining}, pp.
  \bibinfo{pages}{2623--2631}.
\newblock \DOIprefix\doi{10.1145/3292500.3330701}.
\bibitem[{Albelali and Ahmed(2025)}]{albelali2025hidden_leaks}
\bibinfo{author}{Albelali, S.}, \bibinfo{author}{Ahmed, M.},
  \bibinfo{year}{2025}.
\newblock \bibinfo{title}{Hidden leaks in time series forecasting: How data
  leakage affects {LSTM} evaluation across configurations and validation
  strategies}.
\newblock \bibinfo{howpublished}{arXiv preprint arXiv:2512.06932}.
\bibitem[{Chen and Guestrin(2016)}]{chen2016xgboost}
\bibinfo{author}{Chen, T.}, \bibinfo{author}{Guestrin, C.},
  \bibinfo{year}{2016}.
\newblock \bibinfo{title}{{XGBoost}: A scalable tree boosting system}, in:
  \bibinfo{booktitle}{Proceedings of the 22nd ACM SIGKDD International
  Conference on Knowledge Discovery and Data Mining}, pp.
  \bibinfo{pages}{785--794}.
\newblock \DOIprefix\doi{10.1145/2939672.2939785}.
\bibitem[{Chen et~al.(2025)Chen, Liu, Luo and Zeng}]{b5_transformer_foc}
\bibinfo{author}{Chen, X.}, \bibinfo{author}{Liu, X.}, \bibinfo{author}{Luo,
  Y.}, \bibinfo{author}{Zeng, X.}, \bibinfo{year}{2025}.
\newblock \bibinfo{title}{Exploring time-series deep learning models for ship
  fuel consumption prediction}.
\newblock \bibinfo{journal}{Journal of Marine Science and Engineering}
  \bibinfo{volume}{13}, \bibinfo{pages}{2102}.
\newblock \DOIprefix\doi{10.3390/jmse13112102}.
\bibitem[{Chen et~al.(2024)Chen, Sun, Xie, Li, Li and
  Zhao}]{short_term_foc2024}
\bibinfo{author}{Chen, Y.}, \bibinfo{author}{Sun, B.}, \bibinfo{author}{Xie,
  X.}, \bibinfo{author}{Li, X.}, \bibinfo{author}{Li, Y.},
  \bibinfo{author}{Zhao, Y.}, \bibinfo{year}{2024}.
\newblock \bibinfo{title}{Short-term forecasting for ship fuel consumption
  based on deep learning}.
\newblock \bibinfo{journal}{Ocean Engineering} \bibinfo{volume}{301},
  \bibinfo{pages}{117398}.
\newblock \DOIprefix\doi{10.1016/j.oceaneng.2024.117398}.
\bibitem[{Chua et~al.(2025)Chua, Chieng, Wang, Lee, Xu, Fu and
  Yan}]{noise_robustness2025}
\bibinfo{author}{Chua, P.C.}, \bibinfo{author}{Chieng, R.},
  \bibinfo{author}{Wang, R.}, \bibinfo{author}{Lee, K.}, \bibinfo{author}{Xu,
  H.}, \bibinfo{author}{Fu, X.}, \bibinfo{author}{Yan, R.},
  \bibinfo{year}{2025}.
\newblock \bibinfo{title}{Development of a knowledge-based noise test for
  robustness of vessel fuel consumption prediction models}.
\newblock \bibinfo{journal}{Maritime Policy \& Management}
  \bibinfo{volume}{52}, \bibinfo{pages}{1178--1207}.
\newblock \DOIprefix\doi{10.1080/03088839.2025.2548786}.
\bibitem[{Fan et~al.(2026)Fan, Sun, Hu, Vladimir and
  Mao}]{a2_transfer_scarcity}
\bibinfo{author}{Fan, A.}, \bibinfo{author}{Sun, S.}, \bibinfo{author}{Hu, Z.},
  \bibinfo{author}{Vladimir, N.}, \bibinfo{author}{Mao, W.},
  \bibinfo{year}{2026}.
\newblock \bibinfo{title}{Enhancing transfer learning strategies for ship fuel
  consumption prediction under data scarcity}.
\newblock \bibinfo{journal}{Ocean Engineering} \bibinfo{volume}{351},
  \bibinfo{pages}{124398}.
\newblock \DOIprefix\doi{10.1016/j.oceaneng.2026.124398}.
\bibitem[{Fan et~al.(2024a)Fan, Wang, Yang, Tu, Yang and
  Shu}]{l4_comprehensive_eval}
\bibinfo{author}{Fan, A.}, \bibinfo{author}{Wang, Y.}, \bibinfo{author}{Yang,
  L.}, \bibinfo{author}{Tu, X.}, \bibinfo{author}{Yang, J.},
  \bibinfo{author}{Shu, Y.}, \bibinfo{year}{2024}a.
\newblock \bibinfo{title}{Comprehensive evaluation of machine learning models
  for predicting ship energy consumption based on onboard sensor data}.
\newblock \bibinfo{journal}{Ocean and Coastal Management}
  \bibinfo{volume}{248}, \bibinfo{pages}{106946}.
\newblock \DOIprefix\doi{10.1016/j.ocecoaman.2023.106946}.
\bibitem[{Fan et~al.(2024b)Fan, Chen, Mo and Liao}]{k2_federated_building}
\bibinfo{author}{Fan, C.}, \bibinfo{author}{Chen, R.}, \bibinfo{author}{Mo,
  J.}, \bibinfo{author}{Liao, L.}, \bibinfo{year}{2024}b.
\newblock \bibinfo{title}{Personalized federated learning for cross-building
  energy knowledge sharing: Cost-effective strategies and model architectures}.
\newblock \bibinfo{journal}{Applied Energy} \bibinfo{volume}{362},
  \bibinfo{pages}{123016}.
\newblock \DOIprefix\doi{10.1016/j.apenergy.2024.123016}.
\bibitem[{Ghosh et~al.(2026)Ghosh, Sharma and Mattoo}]{ghosh2026unified}
\bibinfo{author}{Ghosh, D.}, \bibinfo{author}{Sharma, R.},
  \bibinfo{author}{Mattoo, A.}, \bibinfo{year}{2026}.
\newblock \bibinfo{title}{A physics-informed unified fleet model for
  sustainable fuel and carbon prediction in maritime shipping}.
\newblock \bibinfo{journal}{Franklin Open} \bibinfo{volume}{15},
  \bibinfo{pages}{100560}.
\newblock \DOIprefix\doi{10.1016/j.fraope.2026.100560}.
\bibitem[{Giannopoulos et~al.(2024)Giannopoulos, Gkonis, Bithas, Nomikos,
  Kalafatelis and Trakadas}]{giannopoulos2024maritime}
\bibinfo{author}{Giannopoulos, A.}, \bibinfo{author}{Gkonis, P.},
  \bibinfo{author}{Bithas, P.}, \bibinfo{author}{Nomikos, N.},
  \bibinfo{author}{Kalafatelis, A.}, \bibinfo{author}{Trakadas, P.},
  \bibinfo{year}{2024}.
\newblock \bibinfo{title}{Federated learning for maritime environments: Use
  cases, experimental results, and open issues}.
\newblock \bibinfo{journal}{Journal of Marine Science and Engineering}
  \bibinfo{volume}{12}, \bibinfo{pages}{1034}.
\newblock \DOIprefix\doi{10.3390/jmse12061034}.
\bibitem[{Gorishniy et~al.(2025)Gorishniy, Kotelnikov and Babenko}]{k6_tabm}
\bibinfo{author}{Gorishniy, Y.}, \bibinfo{author}{Kotelnikov, A.},
  \bibinfo{author}{Babenko, A.}, \bibinfo{year}{2025}.
\newblock \bibinfo{title}{{TabM}: Advancing tabular deep learning with
  parameter-efficient ensembling}, in: \bibinfo{booktitle}{Proceedings of the
  International Conference on Learning Representations (ICLR)}.
\bibitem[{Guo et~al.(2022)Guo, Liang, Tvete, Brinks and
  Vanem}]{d4_hybrid_physics}
\bibinfo{author}{Guo, B.}, \bibinfo{author}{Liang, Q.}, \bibinfo{author}{Tvete,
  H.A.}, \bibinfo{author}{Brinks, H.}, \bibinfo{author}{Vanem, E.},
  \bibinfo{year}{2022}.
\newblock \bibinfo{title}{Combined machine learning and physics-based models
  for estimating fuel consumption of cargo ships}.
\newblock \bibinfo{journal}{Ocean Engineering} \bibinfo{volume}{255},
  \bibinfo{pages}{111435}.
\newblock \DOIprefix\doi{10.1016/j.oceaneng.2022.111435}.
\bibitem[{Guo and Berkhahn(2016)}]{j2_entity_embeddings}
\bibinfo{author}{Guo, C.}, \bibinfo{author}{Berkhahn, F.},
  \bibinfo{year}{2016}.
\newblock \bibinfo{title}{Entity embeddings of categorical variables}.
\newblock \bibinfo{howpublished}{arXiv preprint arXiv:1604.06737}.
\bibitem[{Han et~al.(2024a)Han, Liu, Li, Sun and Yan}]{a5_federated}
\bibinfo{author}{Han, P.}, \bibinfo{author}{Liu, Z.}, \bibinfo{author}{Li, C.},
  \bibinfo{author}{Sun, Z.}, \bibinfo{author}{Yan, C.}, \bibinfo{year}{2024}a.
\newblock \bibinfo{title}{A novel federated learning-based two-stage approach
  for ship energy consumption optimization considering both shipping data
  security and statistical heterogeneity}.
\newblock \bibinfo{journal}{Energy} \bibinfo{volume}{309},
  \bibinfo{pages}{133150}.
\newblock \DOIprefix\doi{10.1016/j.energy.2024.133150}.
\bibitem[{Han et~al.(2024b)Han, Liu, Sun and Yan}]{han2024pfl}
\bibinfo{author}{Han, P.}, \bibinfo{author}{Liu, Z.}, \bibinfo{author}{Sun,
  Z.}, \bibinfo{author}{Yan, C.}, \bibinfo{year}{2024}b.
\newblock \bibinfo{title}{A novel prediction model for ship fuel consumption
  considering shipping data privacy: An {XGBoost-IGWO-LSTM}-based personalized
  federated learning approach}.
\newblock \bibinfo{journal}{Ocean Engineering} \bibinfo{volume}{302},
  \bibinfo{pages}{117668}.
\newblock \DOIprefix\doi{10.1016/j.oceaneng.2024.117668}.
\bibitem[{Han et~al.(2026)Han, Wang, Liu, Zhong and Yan}]{han2025marl_twin}
\bibinfo{author}{Han, P.}, \bibinfo{author}{Wang, Y.}, \bibinfo{author}{Liu,
  Z.}, \bibinfo{author}{Zhong, Y.}, \bibinfo{author}{Yan, C.},
  \bibinfo{year}{2026}.
\newblock \bibinfo{title}{A tailored multi-agent reinforcement learning based
  approach for enhancing collaborative twin ocean-going ships energy
  efficiency}.
\newblock \bibinfo{journal}{Ocean Engineering} \bibinfo{volume}{347},
  \bibinfo{pages}{124032}.
\newblock \DOIprefix\doi{10.1016/j.oceaneng.2025.124032}.
\bibitem[{Handayani et~al.(2025)Handayani, Kim, Lee and Lee}]{handayani2025foc}
\bibinfo{author}{Handayani, M.P.}, \bibinfo{author}{Kim, D.},
  \bibinfo{author}{Lee, S.}, \bibinfo{author}{Lee, J.}, \bibinfo{year}{2025}.
\newblock \bibinfo{title}{Predictive analysis of fuel oil consumption in
  vessels: interpretable modelling with emphasis on load conditions}.
\newblock \bibinfo{journal}{Maritime Policy \& Management}
  \bibinfo{volume}{52}, \bibinfo{pages}{666--689}.
\newblock \DOIprefix\doi{10.1080/03088839.2025.2451363}.
\bibitem[{Holtrop and Mennen(1982)}]{holtrop1982}
\bibinfo{author}{Holtrop, J.}, \bibinfo{author}{Mennen, G.G.J.},
  \bibinfo{year}{1982}.
\newblock \bibinfo{title}{An approximate power prediction method}.
\newblock \bibinfo{journal}{International Shipbuilding Progress}
  \bibinfo{volume}{29}, \bibinfo{pages}{166--170}.
\newblock \DOIprefix\doi{10.3233/ISP-1982-2933501}.
\bibitem[{{International Maritime Organization}(2021)}]{mepc328}
\bibinfo{author}{{International Maritime Organization}}, \bibinfo{year}{2021}.
\newblock \bibinfo{title}{Resolution {MEPC}.328(76): Amendments to {MARPOL}
  Annex {VI}}.
\newblock \bibinfo{type}{Technical Report}. IMO MEPC 76.
\bibitem[{Kairouz et~al.(2021)Kairouz, McMahan et~al.}]{kairouz2021advances}
\bibinfo{author}{Kairouz, P.}, \bibinfo{author}{McMahan, H.B.}, et~al.,
  \bibinfo{year}{2021}.
\newblock \bibinfo{title}{Advances and open problems in federated learning}.
\newblock \bibinfo{journal}{Foundations and Trends in Machine Learning}
  \bibinfo{volume}{14}, \bibinfo{pages}{1--210}.
\newblock \DOIprefix\doi{10.1561/2200000083}.
\bibitem[{Ke et~al.(2017)Ke, Meng, Finley, Wang, Chen, Ma, Ye and
  Liu}]{ke2017lightgbm}
\bibinfo{author}{Ke, G.}, \bibinfo{author}{Meng, Q.}, \bibinfo{author}{Finley,
  T.}, \bibinfo{author}{Wang, T.}, \bibinfo{author}{Chen, W.},
  \bibinfo{author}{Ma, W.}, \bibinfo{author}{Ye, Q.}, \bibinfo{author}{Liu,
  T.Y.}, \bibinfo{year}{2017}.
\newblock \bibinfo{title}{{LightGBM}: A highly efficient gradient boosting
  decision tree}, in: \bibinfo{booktitle}{Advances in Neural Information
  Processing Systems (NeurIPS)}, pp. \bibinfo{pages}{3146--3154}.
\bibitem[{Kim et~al.(2020)Kim, Lee and Lee}]{kim2020svr}
\bibinfo{author}{Kim, D.}, \bibinfo{author}{Lee, S.}, \bibinfo{author}{Lee,
  J.}, \bibinfo{year}{2020}.
\newblock \bibinfo{title}{Data-driven prediction of vessel propulsion power
  using support vector regression with onboard measurement and ocean data}.
\newblock \bibinfo{journal}{Sensors} \bibinfo{volume}{20},
  \bibinfo{pages}{1588}.
\newblock \DOIprefix\doi{10.3390/s20061588}.
\bibitem[{Li et~al.(2024)Li, Wu, Yan, Fang, Li, Gao, Xu and
  Wang}]{k1_cross_building}
\bibinfo{author}{Li, G.}, \bibinfo{author}{Wu, Y.}, \bibinfo{author}{Yan, C.},
  \bibinfo{author}{Fang, X.}, \bibinfo{author}{Li, T.}, \bibinfo{author}{Gao,
  J.}, \bibinfo{author}{Xu, C.}, \bibinfo{author}{Wang, Z.},
  \bibinfo{year}{2024}.
\newblock \bibinfo{title}{An improved transfer learning strategy for short-term
  cross-building energy prediction using data incremental}.
\newblock \bibinfo{journal}{Building Simulation} \bibinfo{volume}{17},
  \bibinfo{pages}{165--183}.
\newblock \DOIprefix\doi{10.1007/s12273-023-1053-x}.
\bibitem[{Li et~al.(2020)Li, Sahu, Zaheer, Sanjabi, Talwalkar and
  Smith}]{li2020fedprox}
\bibinfo{author}{Li, T.}, \bibinfo{author}{Sahu, A.K.},
  \bibinfo{author}{Zaheer, M.}, \bibinfo{author}{Sanjabi, M.},
  \bibinfo{author}{Talwalkar, A.}, \bibinfo{author}{Smith, V.},
  \bibinfo{year}{2020}.
\newblock \bibinfo{title}{Federated optimization in heterogeneous networks},
  in: \bibinfo{booktitle}{Proceedings of Machine Learning and Systems}, pp.
  \bibinfo{pages}{429--450}.
\bibitem[{Li et~al.(2021)Li, Jiang, Zhang, Kamp and Dou}]{li2021fedbn}
\bibinfo{author}{Li, X.}, \bibinfo{author}{Jiang, M.}, \bibinfo{author}{Zhang,
  X.}, \bibinfo{author}{Kamp, M.}, \bibinfo{author}{Dou, Q.},
  \bibinfo{year}{2021}.
\newblock \bibinfo{title}{{FedBN}: Federated learning on non-{IID} features via
  local batch normalization}, in: \bibinfo{booktitle}{International Conference
  on Learning Representations (ICLR)}.
\bibitem[{Lundberg and Lee(2017)}]{lundberg2017shap}
\bibinfo{author}{Lundberg, S.M.}, \bibinfo{author}{Lee, S.I.},
  \bibinfo{year}{2017}.
\newblock \bibinfo{title}{A unified approach to interpreting model
  predictions}, in: \bibinfo{booktitle}{Advances in Neural Information
  Processing Systems (NeurIPS)}, pp. \bibinfo{pages}{4765--4774}.
\bibitem[{Luo et~al.(2024)Luo, Yan, Xu and Wang}]{b2_comparison}
\bibinfo{author}{Luo, X.}, \bibinfo{author}{Yan, R.}, \bibinfo{author}{Xu, L.},
  \bibinfo{author}{Wang, S.}, \bibinfo{year}{2024}.
\newblock \bibinfo{title}{Accuracy and applicability of ship's fuel consumption
  prediction models: A comprehensive comparative analysis}.
\newblock \bibinfo{journal}{Energy} \bibinfo{volume}{310},
  \bibinfo{pages}{133187}.
\newblock \DOIprefix\doi{10.1016/j.energy.2024.133187}.
\bibitem[{Luo et~al.(2025)Luo, Zhang, Han, Yan and Wang}]{a1_ann_transfer}
\bibinfo{author}{Luo, X.}, \bibinfo{author}{Zhang, M.}, \bibinfo{author}{Han,
  Y.}, \bibinfo{author}{Yan, R.}, \bibinfo{author}{Wang, S.},
  \bibinfo{year}{2025}.
\newblock \bibinfo{title}{Ship fuel consumption prediction based on transfer
  learning: Models and applications}.
\newblock \bibinfo{journal}{Engineering Applications of Artificial
  Intelligence} \bibinfo{volume}{141}, \bibinfo{pages}{109769}.
\newblock \DOIprefix\doi{10.1016/j.engappai.2024.109769}.
\bibitem[{Mahmoodi et~al.(2025)Mahmoodi, B{\"o}ling and
  Vettor}]{mahmoodi2025weather}
\bibinfo{author}{Mahmoodi, K.}, \bibinfo{author}{B{\"o}ling, J.},
  \bibinfo{author}{Vettor, R.}, \bibinfo{year}{2025}.
\newblock \bibinfo{title}{Evaluating the influence of marine weather parameters
  uncertainties on the ship fuel consumption with {Monte Carlo} analysis}.
\newblock \bibinfo{journal}{Ocean Engineering} \bibinfo{volume}{341},
  \bibinfo{pages}{122531}.
\newblock \DOIprefix\doi{10.1016/j.oceaneng.2025.122531}.
\bibitem[{Marijan et~al.(2026)Marijan, Mohammed and Zaman}]{i4_survey}
\bibinfo{author}{Marijan, D.}, \bibinfo{author}{Mohammed, H.H.},
  \bibinfo{author}{Zaman, B.}, \bibinfo{year}{2026}.
\newblock \bibinfo{title}{Estimation and optimization of ship fuel consumption
  in maritime: Review, challenges, and future directions}.
\newblock \bibinfo{journal}{Journal of Marine Science and Technology}
  \bibinfo{volume}{31}, \bibinfo{pages}{54--76}.
\newblock \DOIprefix\doi{10.1007/s00773-025-01104-9}.
\bibitem[{McMahan et~al.(2017)McMahan, Moore, Ramage, Hampson and Ag{\"u}era~y
  Arcas}]{mcmahan2017fedavg}
\bibinfo{author}{McMahan, B.}, \bibinfo{author}{Moore, E.},
  \bibinfo{author}{Ramage, D.}, \bibinfo{author}{Hampson, S.},
  \bibinfo{author}{Ag{\"u}era~y Arcas, B.}, \bibinfo{year}{2017}.
\newblock \bibinfo{title}{Communication-efficient learning of deep networks
  from decentralized data}, in: \bibinfo{booktitle}{Proceedings of the 20th
  International Conference on Artificial Intelligence and Statistics
  (AISTATS)}, pp. \bibinfo{pages}{1273--1282}.
\bibitem[{Mohamed et~al.(2026)Mohamed, Hu and
  Hendricks}]{mohamed2026datafusion}
\bibinfo{author}{Mohamed, A.}, \bibinfo{author}{Hu, X.},
  \bibinfo{author}{Hendricks, C.}, \bibinfo{year}{2026}.
\newblock \bibinfo{title}{Data fusion and machine learning for ship fuel
  consumption modelling -- a case of bulk carrier vessel}.
\newblock \bibinfo{journal}{Ocean Engineering} \bibinfo{volume}{348},
  \bibinfo{pages}{124021}.
\newblock \DOIprefix\doi{10.1016/j.oceaneng.2025.124021}.
\bibitem[{Mohammed et~al.(2025)Mohammed, Marijan and Maressa}]{d1_pikan}
\bibinfo{author}{Mohammed, H.H.}, \bibinfo{author}{Marijan, D.},
  \bibinfo{author}{Maressa, A.}, \bibinfo{year}{2025}.
\newblock \bibinfo{title}{Physics-informed machine learning for vessel shaft
  power and fuel consumption prediction: Interpretable {KAN}-based approach},
  in: \bibinfo{booktitle}{2025 IEEE International Conference on Big Data
  (BigData)}, pp. \bibinfo{pages}{4653--4662}.
\newblock \DOIprefix\doi{10.1109/BigData66926.2025.11400742}.
\bibitem[{Nguyen et~al.(2023)Nguyen, Fu, Ogawa and Zheng}]{nguyen2023multiship}
\bibinfo{author}{Nguyen, S.}, \bibinfo{author}{Fu, X.}, \bibinfo{author}{Ogawa,
  D.}, \bibinfo{author}{Zheng, Q.}, \bibinfo{year}{2023}.
\newblock \bibinfo{title}{An application-oriented testing regime and multi-ship
  predictive modeling for vessel fuel consumption prediction}.
\newblock \bibinfo{journal}{Transportation Research Part E: Logistics and
  Transportation Review} \bibinfo{volume}{177}, \bibinfo{pages}{103261}.
\newblock \DOIprefix\doi{10.1016/j.tre.2023.103261}.
\bibitem[{Pan and Yang(2010)}]{pan2010survey}
\bibinfo{author}{Pan, S.J.}, \bibinfo{author}{Yang, Q.}, \bibinfo{year}{2010}.
\newblock \bibinfo{title}{A survey on transfer learning}.
\newblock \bibinfo{journal}{IEEE Transactions on Knowledge and Data
  Engineering} \bibinfo{volume}{22}, \bibinfo{pages}{1345--1359}.
\newblock \DOIprefix\doi{10.1109/TKDE.2009.191}.
\bibitem[{Pan et~al.(2026)Pan, Xiao, Wang and Yan}]{pan2026ppfl}
\bibinfo{author}{Pan, X.}, \bibinfo{author}{Xiao, Y.}, \bibinfo{author}{Wang,
  J.}, \bibinfo{author}{Yan, R.}, \bibinfo{year}{2026}.
\newblock \bibinfo{title}{A physics-informed and personalised federated
  learning framework for ship fuel consumption prediction}.
\newblock \bibinfo{journal}{Transportation Research Part C: Emerging
  Technologies} \bibinfo{volume}{190}, \bibinfo{pages}{105797}.
\newblock \DOIprefix\doi{10.1016/j.trc.2026.105797}.
\bibitem[{Parkes et~al.(2022)Parkes, Savasta, Sobey and
  Hudson}]{parkes2022fleet}
\bibinfo{author}{Parkes, A.I.}, \bibinfo{author}{Savasta, T.D.},
  \bibinfo{author}{Sobey, A.J.}, \bibinfo{author}{Hudson, D.A.},
  \bibinfo{year}{2022}.
\newblock \bibinfo{title}{Power prediction for a vessel without recorded data
  using data fusion from a fleet of vessels}.
\newblock \bibinfo{journal}{Expert Systems with Applications}
  \bibinfo{volume}{187}, \bibinfo{pages}{115971}.
\newblock \DOIprefix\doi{10.1016/j.eswa.2021.115971}.
\bibitem[{Perez et~al.(2018)Perez, Strub, de~Vries, Dumoulin and
  Courville}]{perez2018film}
\bibinfo{author}{Perez, E.}, \bibinfo{author}{Strub, F.},
  \bibinfo{author}{de~Vries, H.}, \bibinfo{author}{Dumoulin, V.},
  \bibinfo{author}{Courville, A.}, \bibinfo{year}{2018}.
\newblock \bibinfo{title}{{FiLM}: Visual reasoning with a general conditioning
  layer}, in: \bibinfo{booktitle}{Proceedings of the AAAI Conference on
  Artificial Intelligence}.
\newblock \DOIprefix\doi{10.1609/aaai.v32i1.11671}.
\bibitem[{Prokhorenkova et~al.(2018)Prokhorenkova, Gusev, Vorobev, Dorogush and
  Gulin}]{prokhorenkova2018catboost}
\bibinfo{author}{Prokhorenkova, L.}, \bibinfo{author}{Gusev, G.},
  \bibinfo{author}{Vorobev, A.}, \bibinfo{author}{Dorogush, A.V.},
  \bibinfo{author}{Gulin, A.}, \bibinfo{year}{2018}.
\newblock \bibinfo{title}{{CatBoost}: Unbiased boosting with categorical
  features}, in: \bibinfo{booktitle}{Advances in Neural Information Processing
  Systems (NeurIPS)}.
\bibitem[{Rosenblatt et~al.(2024)Rosenblatt, Tejavibulya, Jiang, Noble and
  Scheinost}]{rosenblatt2024leakage}
\bibinfo{author}{Rosenblatt, M.}, \bibinfo{author}{Tejavibulya, L.},
  \bibinfo{author}{Jiang, R.}, \bibinfo{author}{Noble, S.},
  \bibinfo{author}{Scheinost, D.}, \bibinfo{year}{2024}.
\newblock \bibinfo{title}{Data leakage inflates prediction performance in
  connectome-based machine learning models}.
\newblock \bibinfo{journal}{Nature Communications} \bibinfo{volume}{15},
  \bibinfo{pages}{1829}.
\newblock \DOIprefix\doi{10.1038/s41467-024-46150-w}.
\bibitem[{Schultz(2007)}]{schultz2007}
\bibinfo{author}{Schultz, M.P.}, \bibinfo{year}{2007}.
\newblock \bibinfo{title}{Effects of coating roughness and biofouling on ship
  resistance and powering}.
\newblock \bibinfo{journal}{Biofouling} \bibinfo{volume}{23},
  \bibinfo{pages}{331--341}.
\newblock \DOIprefix\doi{10.1080/08927010701461974}.
\bibitem[{Son and Kim(2026)}]{son2025varnn}
\bibinfo{author}{Son, J.}, \bibinfo{author}{Kim, J.H.}, \bibinfo{year}{2026}.
\newblock \bibinfo{title}{Pre-departure ship fuel consumption prediction under
  out-of-distribution condition}.
\newblock \bibinfo{journal}{Expert Systems with Applications}
  \bibinfo{volume}{305}, \bibinfo{pages}{130883}.
\newblock \DOIprefix\doi{10.1016/j.eswa.2025.130883}.
\bibitem[{Viga et~al.(2026)Viga, Mueck, L{\"o}ser and Weis}]{b1_fuelcast}
\bibinfo{author}{Viga, J.}, \bibinfo{author}{Mueck, P.},
  \bibinfo{author}{L{\"o}ser, A.}, \bibinfo{author}{Weis, T.},
  \bibinfo{year}{2026}.
\newblock \bibinfo{title}{{FuelCast}: Benchmarking tabular and temporal models
  for ship fuel consumption}, in: \bibinfo{booktitle}{Advanced Analytics and
  Learning on Temporal Data (AALTD 2025)}, \bibinfo{publisher}{Springer},
  \bibinfo{address}{Cham}. pp. \bibinfo{pages}{54--69}.
\newblock \DOIprefix\doi{10.1007/978-3-032-15535-1_4}.
\bibitem[{Vorkapi{\'c}(2026)}]{vorkapic2026chronological}
\bibinfo{author}{Vorkapi{\'c}, A.}, \bibinfo{year}{2026}.
\newblock \bibinfo{title}{Toward realistic ship fuel consumption prediction
  under chronological validation}.
\newblock \bibinfo{journal}{Journal of Marine Science and Engineering}
  \bibinfo{volume}{14}, \bibinfo{pages}{538}.
\newblock \DOIprefix\doi{10.3390/jmse14060538}.
\bibitem[{Wang et~al.(2023a)Wang, Yan, Au, Wang and Jin}]{federated_green2023}
\bibinfo{author}{Wang, H.}, \bibinfo{author}{Yan, R.}, \bibinfo{author}{Au,
  M.H.}, \bibinfo{author}{Wang, S.}, \bibinfo{author}{Jin, Y.J.},
  \bibinfo{year}{2023}a.
\newblock \bibinfo{title}{Federated learning for green shipping optimization
  and management}.
\newblock \bibinfo{journal}{Advanced Engineering Informatics}
  \bibinfo{volume}{56}, \bibinfo{pages}{101994}.
\newblock \DOIprefix\doi{10.1016/j.aei.2023.101994}.
\bibitem[{Wang et~al.(2023b)Wang, Hua, Huang, Guo, Liu, Ma, Ma and
  Jiang}]{wang2023ga_lstm}
\bibinfo{author}{Wang, K.}, \bibinfo{author}{Hua, Y.}, \bibinfo{author}{Huang,
  L.}, \bibinfo{author}{Guo, X.}, \bibinfo{author}{Liu, X.},
  \bibinfo{author}{Ma, Z.}, \bibinfo{author}{Ma, R.}, \bibinfo{author}{Jiang,
  X.}, \bibinfo{year}{2023}b.
\newblock \bibinfo{title}{A novel {GA-LSTM}-based prediction method of ship
  energy usage based on the characteristics analysis of operational data}.
\newblock \bibinfo{journal}{Energy} \bibinfo{volume}{282},
  \bibinfo{pages}{128910}.
\newblock \DOIprefix\doi{10.1016/j.energy.2023.128910}.
\bibitem[{Wang et~al.(2025)Wang, Wang, Ren, Chen and Yan}]{ftl_sm2025}
\bibinfo{author}{Wang, S.}, \bibinfo{author}{Wang, R.}, \bibinfo{author}{Ren,
  F.}, \bibinfo{author}{Chen, M.}, \bibinfo{author}{Yan, R.},
  \bibinfo{year}{2025}.
\newblock \bibinfo{title}{A privacy-preserving federated transfer learning with
  ship mapping ({FTL-SM}) framework for accurate ship energy efficiency
  prediction}.
\newblock \bibinfo{journal}{Advanced Engineering Informatics}
  \bibinfo{volume}{68}, \bibinfo{pages}{103569}.
\newblock \DOIprefix\doi{10.1016/j.aei.2025.103569}.
\bibitem[{Yosinski et~al.(2014)Yosinski, Clune, Bengio and
  Lipson}]{yosinski2014transferable}
\bibinfo{author}{Yosinski, J.}, \bibinfo{author}{Clune, J.},
  \bibinfo{author}{Bengio, Y.}, \bibinfo{author}{Lipson, H.},
  \bibinfo{year}{2014}.
\newblock \bibinfo{title}{How transferable are features in deep neural
  networks?}, in: \bibinfo{booktitle}{Advances in Neural Information Processing
  Systems}, pp. \bibinfo{pages}{3320--3328}.
\bibitem[{Zhang et~al.(2024)Zhang, Tsoulakos, Kujala and
  Hirdaris}]{d5_cnn_lstm}
\bibinfo{author}{Zhang, M.}, \bibinfo{author}{Tsoulakos, N.},
  \bibinfo{author}{Kujala, P.}, \bibinfo{author}{Hirdaris, S.},
  \bibinfo{year}{2024}.
\newblock \bibinfo{title}{A deep learning method for the prediction of ship
  fuel consumption in real operational conditions}.
\newblock \bibinfo{journal}{Engineering Applications of Artificial
  Intelligence} \bibinfo{volume}{130}, \bibinfo{pages}{107425}.
\newblock \DOIprefix\doi{10.1016/j.engappai.2023.107425}.
\bibitem[{Zhao and Shen(2025)}]{f2_proactive_drift}
\bibinfo{author}{Zhao, L.}, \bibinfo{author}{Shen, Y.}, \bibinfo{year}{2025}.
\newblock \bibinfo{title}{Proactive model adaptation against concept drift for
  online time series forecasting}, in: \bibinfo{booktitle}{Proceedings of the
  31st ACM SIGKDD Conference on Knowledge Discovery and Data Mining}, pp.
  \bibinfo{pages}{2020--2031}.
\newblock \DOIprefix\doi{10.1145/3690624.3709210}.
\bibitem[{Zhao et~al.(2026)Zhao, Liu and Prakash}]{f1_concept_drift}
\bibinfo{author}{Zhao, Z.}, \bibinfo{author}{Liu, H.},
  \bibinfo{author}{Prakash, B.A.}, \bibinfo{year}{2026}.
\newblock \bibinfo{title}{Tackling time-series forecasting generalization via
  mitigating concept drift}, in: \bibinfo{booktitle}{International Conference
  on Learning Representations (ICLR)}.
\bibitem[{Zhou et~al.(2027)Zhou, Jin, Zhang and Yan}]{zhou2026f2vformer}
\bibinfo{author}{Zhou, L.}, \bibinfo{author}{Jin, J.G.},
  \bibinfo{author}{Zhang, Y.}, \bibinfo{author}{Yan, R.}, \bibinfo{year}{2027}.
\newblock \bibinfo{title}{{F2V-Former}: Fleet-to-vessel transfer learning
  guided by static attributes for container vessel fuel consumption
  prediction}.
\newblock \bibinfo{journal}{Transportation Research Part C: Emerging
  Technologies} \bibinfo{volume}{194}, \bibinfo{pages}{105986}.
\newblock \DOIprefix\doi{10.1016/j.trc.2026.105986}.

\end{thebibliography}

\appendix
\section{Implementation Details}
\label{app:implementation}

This appendix provides the implementation details for all 11 methods. All neural networks are implemented in PyTorch; tree-based models use LightGBM 4.6 or CatBoost 1.2. Neural networks use StandardScaler fitted without test observations; tree-based models operate on raw features. Global and Per-Ship ANN scaling uses the training split, while the federated implementations use the combined training and validation pool. All experiments use random seed 42. The ablation (Section~\ref{sec:ablation}) is additionally repeated over five seeds, and the in-fleet and leave-one-ship-out tests of F2V-Former over three seeds.

\subsection{Model Configurations}

Table~\ref{tab:app_config} summarizes the architecture and training configuration for each method.

\begin{table}[htbp]
\centering
\caption{Training configuration for all 11 methods. LR: learning rate; BS: batch size; ES: early stopping patience.}
\label{tab:app_config}
\footnotesize
\begin{tabular}{llllllll}
\toprule
Method & Architecture & Optimizer & LR & Loss & BS & Max iter. & ES \\
\midrule
Global LGBM & GBDT & --- & HPO & Custom MAPE & --- & 3{,}000 & 50 \\
CatBoost & Ordered GBDT & --- & HPO & Native MAPE & --- & 3{,}000 & 50 \\
TabM-FiLM & TabM + FiLM & AdamW & HPO & L1 & HPO & 200 ep. & 25 \\
Global ANN & $[256,128,64]$ + emb(16) & Adam & $10^{-3}$ & L1 & 512 & 200 ep. & 25 \\
ANN Transfer & Global ANN $\to$ FT & Adam & $5\!\times\!10^{-5}$ & L1 & 256 & 30 ep. & 10 \\
F2V-Former & Transformer (4 blocks) & Adam & $3\!\times\!10^{-4}$ & L1 & 128 & 200 ep. & 25 \\
F2V-Former + LoRA & F2V-Former $\to$ LoRA & Adam & $10^{-3}$ & L1 & 128 & 300 ep. & 30 \\
FedAvg & $[256,128,64] \times 21$ & Adam & $10^{-3}$ & L1 & 256 & 50 rnd. & 10 \\
FedAvg + FT & FedAvg $\to$ FT & Adam & $10^{-4}$ & L1 & 256 & 50 ep. & 10 \\
Per-Ship LGBM & GBDT $\times$ 21 & --- & HPO & Custom MAPE & --- & 3{,}000 & 50 \\
Per-Ship ANN & $[128,64,32] \times 21$ & Adam & $10^{-3}$ & L1 & 256 & 200 ep. & 25 \\
\bottomrule
\end{tabular}
\end{table}

\subsection{Hyperparameter Search Spaces}

Table~\ref{tab:app_hpo_tree} and Table~\ref{tab:app_hpo_nn} report the HPO search spaces. All tree-based models use Optuna with TPE sampling; TabM-FiLM uses the same.

\begin{table}[htbp]
\centering
\caption{HPO search space for tree-based models. Learning-rate and regularization ranges are sampled on a log scale; the remaining parameters use their stated integer or fractional bounds. Global and Per-Ship LightGBM share the custom MAPE objective.}
\label{tab:app_hpo_tree}
\footnotesize
\begin{tabular}{llll}
\toprule
Parameter & Global LGBM (30 trials) & CatBoost (30 trials) & Per-Ship LGBM (10 trials) \\
\midrule
num\_leaves / depth & $[31, 255]$ & depth $[4, 10]$ & $[15, 127]$ \\
max\_depth & $[4, 12]$ & --- & $[3, 10]$ \\
Learning rate & $[5\!\times\!10^{-3}, 0.2]$ & $[5\!\times\!10^{-3}, 0.2]$ & $[5\!\times\!10^{-3}, 0.2]$ \\
Subsample & $[0.6, 1.0]$ & $[0.6, 1.0]$ & $[0.5, 1.0]$ \\
Colsample & $[0.6, 1.0]$ & $[0.6, 1.0]$ & $[0.5, 1.0]$ \\
min\_child\_samples & $[10, 100]$ & $[10, 100]$ & adaptive\textsuperscript{$\dagger$} \\
L1 reg. & $[10^{-4}, 10]$ & --- & $[10^{-4}, 10]$ \\
L2 reg. & $[10^{-4}, 10]$ & $[10^{-2}, 10]$ & $[10^{-4}, 10]$ \\
max\_bin & --- & $[128, 512]$ & --- \\
\bottomrule
\end{tabular}
\\[2pt]
{\footnotesize \textsuperscript{$\dagger$}Adaptive: lower bound $= \max(5,\, n_{\text{train}}/100)$, upper bound $= \max(10,\, n_{\text{train}}/10)$.}
\end{table}

\begin{table}[htbp]
\centering
\caption{HPO search space for TabM-FiLM (30 Optuna trials).}
\label{tab:app_hpo_nn}
\footnotesize
\begin{tabular}{ll}
\toprule
Parameter & Range \\
\midrule
Ensemble size $k$ & $\{32, 64\}$ \\
Block dimension & $\{192, 256, 384\}$ \\
Number of blocks & $[2, 4]$ \\
Learning rate & $[5\!\times\!10^{-5},\, 2\!\times\!10^{-3}]$ (log) \\
Dropout & $[0.05, 0.45]$ \\
Weight decay & $[10^{-5}, 10^{-2}]$ (log) \\
Batch size & $\{512, 1024, 2048\}$ \\
\bottomrule
\end{tabular}
\end{table}

\paragraph{Fixed LOSO and ablation settings.} The primary LOSO fits use 500 boosting rounds, 127 leaves, maximum depth 8, learning rate 0.05, feature fraction 0.8, minimum child samples 30, L1 regularization 0.1, L2 regularization 1.0, and seed 42. Row subsampling is inactive in the primary LOSO run (bagging frequency zero). The repeated ablations use seeds 42, 1, 2, 3, and 4, with row fraction 0.8 and bagging frequency 1; the other settings and 500-round duration are fixed. These settings are distinct from the HPO-selected full-fleet LightGBM. The calendar-restricted checks in Section~\ref{sec:results_comparison} refit the HPO-selected configuration (2{,}412 rounds) and the primary LOSO configuration for each target vessel. The bootstrap resamples whole test voyage legs within each vessel. In the few-shot extension of the leave-one-ship-out tests, the pooled refit uses the fixed LOSO configuration. Continued boosting starts from the LOSO model with learning rate 0.02 and selects at most 300 additional rounds (patience 20) on the last 20\% of the $n$ records before refitting on all $n$. The per-ship model uses the fixed configuration, starts from the mean of its records, and selects at most 500 rounds in the same way. Records are taken in time order from the vessel's training pool. The identity-removal configurations (F2$^{-}$, F4$^{-}$) use the same five seeds and settings as the other ablation configurations.

\paragraph{Input perturbations.} The stored Global and Per-Ship LightGBM models are applied to perturbed copies of the test inputs, five draws per scenario. Wave height, wave period, wind speed, and current speed are scaled by $1+\varepsilon$ and truncated at zero; direction noise is added to the wind, wave, and current directions; speed noise is added to speed over ground, with a floor of 0.1\,kn. Per-leg scenarios draw one value per voyage leg and apply it to all of its records. Heading-relative components, speed polynomials, interaction terms, and cross-ship ratios are recomputed from the perturbed inputs.

\subsection{Neural Network Architectures}

Tables~\ref{tab:app_global_ann}--\ref{tab:app_pership_ann} detail the layer-by-layer architecture for each neural network variant. All fully connected blocks follow the same pattern: Linear $\to$ BatchNorm1d $\to$ ReLU $\to$ Dropout(0.2).

\begin{table}[htbp]
\centering
\caption{Global ANN architecture (also used as pretrained base for ANN Transfer). $n_c = 46$ continuous features.}
\label{tab:app_global_ann}
\footnotesize
\begin{tabular}{clccl}
\toprule
\# & Layer & Input dim & Output dim & Notes \\
\midrule
0 & Embedding(21, 16) & ship index & 16 & Learned ship embedding \\
1 & Concatenate & $n_c + 16$ & 62 & Continuous features + embedding \\
2 & FC block & 62 & 256 & Linear + BN + ReLU + Drop(0.2) \\
3 & FC block & 256 & 128 & Linear + BN + ReLU + Drop(0.2) \\
4 & FC block & 128 & 64 & Linear + BN + ReLU + Drop(0.2) \\
5 & Linear & 64 & 1 & Output (FOC, t/h) \\
\midrule
\multicolumn{4}{l}{Total parameters} & 58{,}577 \\
\bottomrule
\end{tabular}
\end{table}

\begin{table}[htbp]
\centering
\caption{FedANN architecture (used by FedAvg and FedAvg+FT). No ship embedding is included in this implementation. $n_c = 46$ continuous features.}
\label{tab:app_fedann}
\footnotesize
\begin{tabular}{clccl}
\toprule
\# & Layer & Input dim & Output dim & Notes \\
\midrule
1 & FC block & $n_c = 46$ & 256 & Linear + BN + ReLU + Drop(0.2) \\
2 & FC block & 256 & 128 & Linear + BN + ReLU + Drop(0.2) \\
3 & FC block & 128 & 64 & Linear + BN + ReLU + Drop(0.2) \\
4 & Linear & 64 & 1 & Output (FOC, t/h) \\
\midrule
\multicolumn{4}{l}{Total parameters} & 54{,}145 \\
\bottomrule
\end{tabular}
\end{table}

\begin{table}[htbp]
\centering
\caption{Per-Ship ANN architecture. One independent model per vessel; no ship embedding needed. $n_c = 46$ continuous features (no ship identity).}
\label{tab:app_pership_ann}
\footnotesize
\begin{tabular}{clccl}
\toprule
\# & Layer & Input dim & Output dim & Notes \\
\midrule
1 & FC block & $n_c = 46$ & 128 & Linear + BN + ReLU + Drop(0.2) \\
2 & FC block & 128 & 64 & Linear + BN + ReLU + Drop(0.2) \\
3 & FC block & 64 & 32 & Linear + BN + ReLU + Drop(0.2) \\
4 & Linear & 32 & 1 & Output (FOC, t/h) \\
\midrule
\multicolumn{4}{l}{Total parameters} & 16{,}833 \\
\bottomrule
\end{tabular}
\end{table}

\subsection{Training Protocols}

\textbf{Model selection and final fitting.} The non-federated models select hyperparameters or training duration on the fixed validation partition and are then fitted on the combined training and validation pool. ANN Transfer starts from the resulting full-pool Global ANN. FedAvg first selects a checkpoint by validation MAPE, then continues training from that checkpoint on the combined pool for the selected number of additional rounds. FedAvg+FT instead starts from the validation-selected federated checkpoint and performs target-vessel fine-tuning. Test records are excluded throughout. Thus these implementations share the test protocol but do not use identical final-fitting procedures.

\textbf{ANN Transfer fine-tuning.} The pooled Global ANN initializes each target-vessel model. Its first two fully connected blocks are frozen; the final hidden block, output layer, and ship embedding are trainable. Fine-tuning uses Adam with learning rate $5\times10^{-5}$, at most 30 epochs, and patience 10. Validation compares each epoch with the unchanged initial model, allowing zero fine-tuning epochs. The selected epoch count is then applied from the same initialization on the target vessel's combined training and validation records. Those validation records have already contributed to the pretrained model, so this validation is not independent of the pretraining.

\textbf{F2V-Former.} The re-implementation follows the architecture of \citet{zhou2026f2vformer}. A static encoder embeds the seven ship specifications, and a dynamic encoder with sinusoidal positional encoding embeds the other 39 inputs of the record and of its seven preceding records in the same voyage leg. Four Transformer blocks (width 256, eight heads) use conditional layer normalization guided by the static encoding and cross-attention to the specifications grouped into three tokens, namely ship age, the hull dimensions (deadweight, length, and breadth), and the propulsion (engine rating and propeller diameter and pitch). Mean pooling and a two-layer head give the prediction, with 4.61 million parameters in total. Windows at the start of a leg are padded and masked so that every test record is predicted, and FOC never enters a window. The deviations from the original study follow from the inputs available before departure. Since no route attributes exist, the specifications form three instead of four static tokens, and the window of 14 daily noon reports is replaced by the 3-hour records of the same leg. The window length, depth, width, learning rate, and loss were selected from 13 configurations on the validation legs. The training details that the original study does not report follow the Global ANN, namely Adam with weight decay $10^{-5}$, learning-rate reduction on plateaus, early stopping on the validation legs, and refitting on the combined training and validation records for the selected number of epochs with a cosine schedule. F2V-Former + LoRA trains rank-16 updates ($\alpha = 32$) of all 51 linear layers on the records of each vessel, with the learning rate selected from $\{10^{-4}, 3\times10^{-4}, 10^{-3}\}$ on the validation legs. Its number of epochs is selected on the validation legs of each vessel with the model trained on the training legs only, and zero epochs are allowed. The leave-one-ship-out tests reuse the selected configuration with three seeds. The few-shot tests use one seed and select the number of LoRA epochs on the last 20\% of the $n$ records.

\textbf{FedAvg protocol.} Algorithm~\ref{alg:fedavg} describes validation-based checkpoint selection. Aggregation weights clients by local sample count. The implementations use common scaling statistics calculated from the combined training and validation pool; the benchmark is therefore a federated training simulation with centralized preprocessing, not a complete privacy-preserving deployment. Plain FedAvg subsequently continues from the selected checkpoint on combined training and validation records, whereas FedAvg+FT uses the selected checkpoint directly for personalization.

\begin{algorithm}[htbp]
\caption{Federated Averaging (FedAvg) for FOC Prediction}
\label{alg:fedavg}
\begin{algorithmic}[1]
\State Initialize global model weights $\mathbf{w}_0$
\For{round $r = 1, 2, \ldots, R_{\max}$ ($R_{\max} = 50$)}
    \For{each client $k \in \{1, \ldots, 21\}$}
        \State $\mathbf{w}_k^r \gets$ LocalTrain($\mathbf{w}_{r-1}$, $\mathcal{D}_k$, $E=5$, lr$=10^{-3}$)
    \EndFor
    \State $\mathbf{w}_r \gets \sum_{k=1}^{21} \frac{n_k}{N} \mathbf{w}_k^r$ \Comment{Size-weighted averaging}
    \State Evaluate mean MAPE on all clients' validation sets
    \If{no improvement for 10 rounds}
        \State \textbf{break}
    \EndIf
\EndFor
\State \Return $\mathbf{w}_{r^*}$ at best validation round $r^*$
\end{algorithmic}
\end{algorithm}

\textbf{FedAvg + Fine-Tuning.} Each target model starts from the validation-selected FedANN checkpoint. The implementation freezes all but the final eight parameter tensors: the second hidden block's batch-normalization parameters, the final hidden block's linear and batch-normalization parameters, and the output layer. Fine-tuning uses learning rate $10^{-4}$, at most 50 epochs, and patience 10. Zero fine-tuning epochs are allowed. The selected duration is applied from the same checkpoint on the target vessel's combined training and validation records.

\end{document}